\documentclass[10pt,towcolumn,twoside]{IEEEtran}
\usepackage{amsmath}
\usepackage{amsthm}
\usepackage{algorithmic}
\usepackage{multirow}
\usepackage{bigstrut}
\usepackage{graphicx}
\usepackage{hhline}
\usepackage{cite}
\usepackage{algorithm}
\usepackage{amssymb}
\usepackage{epsfig}
\usepackage{array}
\usepackage{subfigure}
\usepackage{bm}
\usepackage{enumerate}
\usepackage{arydshln}
\usepackage{extarrows}
\usepackage{color}
\usepackage[colorlinks,linkcolor=black,anchorcolor=black,citecolor=black,urlcolor=black]{hyperref}

\newtheorem{theorem}{Theorem}
\newtheorem{lemma}[theorem]{Lemma}

\theoremstyle{definition}

\theoremstyle{remark}


\def\cG{\mathcal{G}}
\def\cN{\mathcal{N}}
\def\cE{\mathcal{E}}
\def\cL{\mathcal{L}}
\def\cV{\mathcal{V}}
\def\A{\mathbf{A}}
\def\B{\mathbf{B}}

\def\D{\mathbf{D}}
\def\L{\mathbf{L}}
\def\P{\mathbf{P}}

\def\U{\mathbf{U}}
\def\V{\mathbf{V}}
\def\H{\mathbf{H}}
\def\I{\mathbf{I}}

\def\W{\mathbf{W}}
\def\Del{\boldsymbol{\Delta}}
\def\Lamb{\boldsymbol{\Lambda}}

\def\Xii{\boldsymbol{\Xi}}
\def\x{\mathbf{x}}
\def\y{\mathbf{y}}

\def\v{\mathbf{v}}
\def\h{\mathbf{h}}

\usepackage{amsmath}
{
   \theoremstyle{plain}
   
}

\begin{document}
%
\title{Robust Semi-Supervised Graph Classifier Learning with Negative Edge Weights}

\author{Gene Cheung,~\IEEEmembership{Senior Member,~IEEE},
Weng-Tai Su,~\IEEEmembership{Student Member,~IEEE},
Yu Mao,~\IEEEmembership{Student Member,~IEEE},
Chia-Wen Lin,~\IEEEmembership{Senior Member,~IEEE}

\begin{small}


\thanks{G. Cheung, Y. Mao are with the National Institute of Informatics, Graduate University for Advanced Studies, Tokyo 101-8430, Japan (e-mail: cheung@nii.ac.jp, vimystic@gmail.com).}

\thanks{W.-T. Su, C.-W. Lin are with National Tsing Hua University, Hsinchu, Taiwan 30013 (e-mail: wengtai2008@hotmail.com, cwlin@ee.nthu.edu.tw).}

\end{small}
}

\maketitle

\begin{abstract}
In a semi-supervised learning scenario, (possibly noisy) partially observed labels are used as input to train a classifier, in order to assign labels to unclassified samples.
In this paper, we construct a complete graph-based binary classifier given only samples' feature vectors and partial labels.
Specifically, we first build appropriate similarity graphs with positive and negative edge weights connecting all samples based on inter-node feature distances.
By viewing a binary classifier as a piecewise constant graph-signal, we cast classifier learning as a signal restoration problem via a classical maximum a posteriori (MAP) formulation.
One unfortunate consequence of negative edge weights is that the graph Laplacian matrix $\L$ can be indefinite, and previously proposed graph-signal smoothness prior $\x^T \L \x$ for candidate signal $\x$ can lead to pathological solutions.
In response, we derive a minimum-norm perturbation matrix $\Del$ that preserves $\L$'s eigen-structure---based on a fast lower-bound computation of $\L$'s smallest negative eigenvalue via a novel application of the Haynsworth inertia additivity formula---so that $\L + \Del$ is positive semi-definite, resulting in a stable signal prior.
Further, instead of forcing a hard binary decision for each sample, we define the notion of generalized smoothness on graphs that promotes ambiguity in the classifier signal.
Finally, we propose an algorithm based on iterative reweighted least squares (IRLS) that solves the posed MAP problem efficiently.
Extensive simulation results show that our proposed algorithm outperforms both SVM variants and previous graph-based classifiers using positive-edge graphs noticeably.
\end{abstract}


%
\IEEEpeerreviewmaketitle

\section{Introduction}
\label{sec:intro}
\begin{figure}

\begin{minipage}[b]{.95\linewidth}
 \centering
 \centerline{\includegraphics[width=7.0cm]{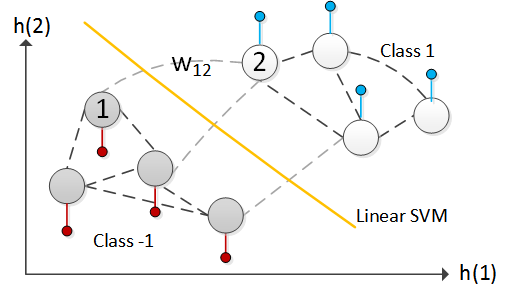}}
\end{minipage}

\vspace{-0.15in}
\caption{Example of a graph classifier and linear SVM in 2-dimensional feature space. The graph $\cG$ contains nodes $\cN$ representing samples, and edges $\cE$ with weights $w_{i,j}$ that reflect feature space distance between nodes. The classifier graph-signal takes on binary values: 1 (blue spikes) and -1 (red spikes).}
\label{fig:graphClass}

\end{figure}

A fundamental problem in machine learning is \textit{semi-supervised learning} \cite{bishop06}: given partially observed labels (possibly corrupted by noise) as input, train a classifier so that unclassified samples can also be appropriately assigned labels.
Among many approaches to the problem is a class of graph-based methods \cite{zhou03,belkin04,gavish10,shuman11,ekambaram13,guillory09,zhang14,chen15,gadde14} that model each sample as a node in a graph, connected to other nodes via undirected edges, with weights that reflect pairwise distances in a high-dimensional feature space. 
See Fig.\;\ref{fig:graphClass} for an example of a graph with eight nodes (samples) in a two-dimensional feature space.
Establishing a graph representation of the data means that intrinsic properties of the graph spectrum (\textit{e.g.}, low graph frequencies that are eigenvectors of the graph Laplacian matrix) can be exploited for label assignment via spectral graph theory \cite{chung96}. 

In this paper, extending previous studies we construct a complete graph-based binary classifier given only samples' feature vectors and partial labels, considering in addition \textit{negative edge weights}.
Conventional formulations in graph signal processing (GSP) \cite{shuman13} use positive edge weights to signify inter-node \textit{similarity}.
However, negative edge weights can signify \textit{dissimilarity}: $w_{i,j}=-1$ means samples $x_i$ and $x_j$ are expected to take on different values, \textit{i.e.} $|x_i - x_j|$ should be large.
Incorporating pairwise dissimilarity into the graph should intuitively be beneficial during classifier learning.
For example, if edge weight $w_{1,2}$ is assigned $-1$ in Fig.\;\ref{fig:graphClass}, then from the graph $\cG$ itself without any label information, one already expects $x_1$ and $x_2$ to be assigned opposite labels in a binary classifier.

Specifically, we first build appropriate similarity graphs with positive and negative edge weights connecting all samples based on inter-node feature distances.
Interpreting a binary classifier as a piecewise constant (PWC) graph-signal, we cast classifier learning as a signal restoration problem via a classical \textit{maximum a posteriori} (MAP) formulation \cite{mao16}.
We show that a graph Laplacian matrix $\L$ with negative edge weights can be indefinite, and a common graph-signal prior called \textit{graph Laplacian regularizer} \cite{hu13, narang13,narang13icassp,mao16tmm,wan16,liu15,hu16spl,pang17} $\x^T \L \x$ for candidate signal $\x$---measuring signal smoothness with respect to the underlying graph---can lead to pathological solutions.
In response, we derive a minimum-norm perturbation matrix $\Del$ that preserves $\L$'s eigen-structure, so that $\L + \Del$ is positive semi-definite (PSD), resulting in a stable signal prior. 
To efficiently compute an approximate $\Del$, we propose a fast recursive algorithm that identifies a lower bound for the smallest eigenvalue of $\L$ via a novel application of the \textit{Haynsworth Inertia Additivity formula} \cite{haynsworth68}.

Second, instead of forcing a hard binary decision for each sample, we define the notion of \textit{generalized smoothness} on graph---an extension of \textit{total generalized variation} (TGV) \cite{bredies15} to the graph-signal domain---that promotes the right amount of ambiguity in the classifier signal. 
Estimated labels with low confidence (signal values close to zero) can be removed thereafter, thus improving the overall classification performance.
We show that by interpreting a graph as an electrical circuit, the generalized smoothness condition is equivalent to the \textit{Kirchhoff's current law} \cite{robbins12}, which helps explain why negative edge weights should not be used when promoting generalized smoothness.

Finally, we propose an algorithm based on \textit{iterative reweighted least squares} (IRLS) \cite{daubechies10} that efficiently solves the posed MAP problem for the noisy label scenario.
Extensive simulation results show that our proposed algorithm outperforms SVM variants, a well-known robust classifier in the machine learning literature called \textit{RobustBoost} \cite{RobustBoost09}, and graph-based classifiers using positive-edge graphs noticeably for both noiseless and noisy label scenarios.

The outline of the paper is as follows. We first overview related works in Section \ref{sec:related}. 
We then review basic GSP concepts and define a graph-signal smoothness prior in Section \ref{sec:prelim}. 
In Section \ref{sec:algo}, we derive an appropriate perturbation matrix $\Del$ such that $\L + \Del$ is PSD.
In Section \ref{sec:fast}, we describe a fast algorithm to approximate the best $\Del$.
In Section \ref{sec:gsmooth}, we introduce the generalized graph-signal smoothness prior, and present an efficient algorithm to solve the MAP problem for classifier learning.  
Finally, we present experimental results and conclusions in Section \ref{sec:results} and \ref{sec:conclude}, respectively.

\section{Related Works}
\label{sec:related}
\subsection{Robust Graph-based Classifier Learning}


There exists a wide range of approaches to noisy label classifier learning, including theoretical (\textit{e.g.}, label propagation in \cite{speriosu11}) and application-specific (\textit{e.g.}, emotion detection using inference algorithm based on multiplicative update rule \cite{wang15}). 
In this paper, we focus specifically on graph-based classifiers, which has been studied extensively in the past decade \cite{zhou03,belkin04,gavish10,shuman11,ekambaram13,guillory09,zhang14,chen15,gadde14}.
An early work \cite{zhou03} used the graph Laplacian matrix to construct an interpolation filter for the input partial labels to compute missing labels. 
Another seminal work \cite{belkin04} proposed two graph-based formulations for the noisy and noiseless label learning scenarios using the graph smoothness prior $\x^{\top} \L \x$. 
While the origin of our MAP formulation can be traced back to \cite{belkin04} and enjoys similar computation benefit of solving simple linear systems at each iteration, neither \cite{zhou03} nor \cite{belkin04} handled the case when negative edges are present to denote inter-node dissimilarity, which is one focus of this paper. 

Recent advent in graph signal processing (GSP) \cite{shuman13} has led to the development of transforms \cite{hu15} and wavelets for signals that live on irregular data kernels described by graphs. 
Using these developed tools, one can design learning algorithms via assumptions in the transform domain \cite{gavish10,shuman11,ekambaram13}.
For example, \cite{ekambaram13} assumed that the $L_1$-norm of the graph wavelet coefficients is small as a signal prior for graph-signal (classifier) reconstruction. 
However, to-date critically sampled perfect reconstruction graph wavelets only exist for very special graphs like bipartite or $k$-colorable graphs. 
Thus, for signals on a general graph, to use these wavelets one must first approximate the original graph with a series of bipartite graphs \cite{zeng16}.
This means that the $L_1$-norm is difficult to apply across different stages of bipartite graph approximation. As a representative graph wavelet scheme, we will show in our experiments that our proposed smoothness prior leads to better performance than an over-complete wavelet in \cite{ekambaram13}.

One can also approach the semi-supervised learning problem from a sampling perspective: available labels are observed signal samples, and missing samples are interpolated using a bandlimited signal assumption \cite{narang13icassp,shomorony14,chen15sampling}.
There are two problems to this approach. 
First, practical graph signals are often not strictly bandlimited. 
Second, observed labels are often corrupted by noise, and straightforward interpolation schemes would lead to error propagation. 
We will show in our experiments that our proposed classifier scheme outperforms \cite{narang13icassp} noticeably.

Compared to previous graph-based classifiers, we make the following three key technical contributions. First, we construct a similarity graph with positive and negative edge weights, latter of which signify inter-node dissimilarities, given only the samples' feature vectors.
Second, for the graph-signal smoothness prior to be numerically stable in a classical MAP formulation, we derive a minimum-norm perturbation matrix $\Del$ that preserves the eigen-structure of the original Laplacian $\L$, so that $\L + \Del$ is PSD via a novel application of the Haynsworth inertia additivity formula \cite{haynsworth68}. 
Third, we extend the generalized smoothness notion in TGV \cite{bredies15} to the graph-signal domain---which we interpret intuitively as Kirchoff's current law \cite{robbins12}---to promote appropriate degree of ambiguity in the classifier solution to lower overall classification error rate.

\subsection{Graph-Signal Image Restoration}

More generally, graph-signal priors have been used for image restoration problems such as denoising  \cite{hu13,pang17}, interpolation \cite{narang13,narang13icassp,mao16tmm}, bit-depth enhancement \cite{wan16} and JPEG de-quantization \cite{liu15,hu16spl}. 
The common assumption is that the desired graph-signal is smooth or bandlimited with respect to an appropriate graph with positive edge weights that reflect inter-pixel similarity. 
Instead of posing an optimization, graph filters can also be designed directly for image denoising \cite{knyazev15_mlsp}, edge-enhancing \cite{knyazev15} and image magnification \cite{gadde15_globalSIP}. 
In contrast, by introducing negative edges into the graph, we incorporate dissimilarity information into a classical MAP formulation like \cite{belkin04} and study methods to resolve the graph Laplacian's indefiniteness. 
Further, we define a generalized notion of graph smoothness for signal restoration specifically for classifier learning. 


\subsection{Negative Edge Weights in Graphs}

Recent studies in the control community have examined the conditions where one or more negative edge weights would induce a graph Laplacian to be indefinite \cite{zelazo14,chen16}.
The analysis, however, rests on an assumption that there are no cycles in the graph with more than one negative edge, which is too restrictive for binary classifier graphs.

\cite{chu16} considered a signed social network where each edge denotes either a cohesive (positive edge weight) or oppositive (negative edge weight) relationship between two vertices. 
The goal is to identify similar groups within the graph, and thus is akin to a distributed clustering problem, which is unsupervised by definition.
In contrast, our goal is to restore a classifier graph-signal from partially observed labels, which is a semi-supervised learning problem.

A notable recent work \cite{knyazev17arxiv} argued that the eigenvectors of the original graph Laplacian are more intuitive and useful than the eigenvectors of the signed graph Laplacian \cite{kunegis10} for spectral clustering. 
The key argument is that the shapes of the first eigenvectors of the original graph Laplacian are more pronounced at the negative edge endpoints, and the condition numbers are more favorable.
In Section\;\ref{sec:algo}, we also stress the usefulness of the eigenvectors of the original graph Laplacian, but are using them for classifier learning rather than clustering.

\section{Preliminaries}
\label{sec:prelim}
\subsection{Graph Definition}
\label{subsec:defn}

We first introduce definitions in GSP needed to formulate our problem. 
A graph $\mathcal{G}(\cV, \mathcal{E}, \mathbf{W})$ has a set $\cV$ of $N$ nodes and a set $\mathcal{E}$ of $M$ edges. 
Each edge $(i,j) \in \mathcal{E}$ connecting nodes $i$ and $j$ is undirected and has an associated scalar weight $w_{i,j}$. 
In this paper, we assume that $w_{i,j}$ can be positive \textit{or} negative; a negative $w_{i,j}$ means that the samples in nodes $i$ and $j$ are \textit{dissimilar}---the samples are expected to have very different values.

A graph-signal $\mathbf{x} \in \mathbb{R}^N$ on $\mathcal{G}$ is a discrete signal of dimension $N$---one value $x_i$ for each node (sample) $i$ in $\mathcal{V}$.
If we restrict $\x$ to be a binary classifier, then $x_i$ can only take on one of two values specifying the class to which sample $i$ belongs, \textit{i.e.}, $\x \in \{-1, 1\}^N$. 
However, letting the reconstructed signal $\hat{\x}$ take on real values $\mathbb{R}^N$ allows us to introduce ambiguity in the reconstruction instead of forcing hard binary decisions; this is discussed in Section\;\ref{sec:gsmooth}.

\subsection{Graph Spectrum}
\label{subsec:spectrum}

Given edge weight (adjacency) matrix $\mathbf{W}$, we define a diagonal \textit{degree matrix} $\mathbf{D}$, where $d_{i,i} = \sum_{j} w_{i,j}$. 
A \textit{combinatorial graph Laplacian matrix} $\mathbf{L}$ is simply $\mathbf{L} = \mathbf{D} - \mathbf{W}$ \cite{shuman13}.
$\mathbf{L}$ is symmetric, which means that it can be eigen-decomposed into:
\begin{equation}
\L = \V \Lamb \V^T
\label{eq:spectrum}
\end{equation}
where $\Lamb$ is a diagonal matrix containing real eigenvalues $\lambda_k$ (not necessarily unique), and $\V$ is an eigen-matrix composed of orthogonal eigenvectors $\v_i$ as columns. 
If edge weights $w_{i,j}$ are strictly positive, then one can show that $\mathbf{L}$ is PSD, meaning that $\lambda_k \geq 0, \forall k$ and $\x^T \L \x \geq 0, ~\forall \x$. 
Non-negative eigenvalues $\lambda_k$ can be interpreted as \textit{graph frequencies}, and eigenvectors $\v_k$ interpreted as corresponding graph frequency components. Together they define the \textit{graph spectrum} for graph $\cG$.

In this paper, we consider also negative edge weights $w_{i,j} < 0$, and thus eigenvalues $\lambda_k$ can be negative and $\mathbf{L}$ can be indefinite.
It is then hard to interpret $\L$'s eigenvalues $\lambda_k$ as frequencies, and in general, it is desirable to have a variational operator that is PSD.
Thus it is desirable to add a \textit{perturbation matrix} $\Del$ to $\L$ such that the resultant $\L + \Del$ is PSD. 
We address this problem of finding an optimal $\Del$ in Section \ref{sec:algo}.

\subsection{Graph-Signal Smoothness Prior for Positive Graphs}
\label{subsec:smooth}

\begin{figure}

\begin{minipage}[b]{.24\linewidth}
 \centering
 \centerline{\includegraphics[width=2.5cm]{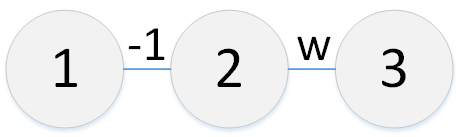}}
 \vspace{0.2cm}
 \centerline{(a) 3-node graph}\medskip
\end{minipage}
\hfill
\begin{minipage}[b]{0.36\linewidth}
\begin{footnotesize}
\begin{equation}
\W \!\! = \!\! \left[ \begin{array}{ccc}
0  & -1 & 0 \\
-1 & 0  & w  \\
0 & w & 0  \\
\end{array}
\right]
\nonumber
\end{equation}
\end{footnotesize}
 \centerline{(b) adjacency $\W$}\medskip
\end{minipage}
\hfill
\begin{minipage}[b]{0.36\linewidth}
\begin{footnotesize}
\begin{equation}
\L \!\! = \!\! \left[ \begin{array}{ccc}
-1  & 1 & 0 \\
1 & w-1  & -w  \\
0 & -w & w  \\
\end{array}
\right]
\nonumber
\end{equation}
\end{footnotesize}
 \centerline{(c) graph Laplacian $\L$}\medskip
\end{minipage}

\vspace{-0.1in}
\caption{Example of a 3-node graph with negative edges.}
\label{fig:graphEx1}

\end{figure}

Traditionally, for graph $\cG$ with positive edge weights, signal $\x$ is considered \textit{smooth} if each sample value $x_i$ on node $i$ is similar to $x_j$ on neighboring nodes $j$ with large $w_{i,j}$.  
In the graph frequency domain, smoothness means that $\mathbf{x}$ contains mostly low graph frequency components; \textit{i.e.}, coefficients $\boldsymbol{\alpha} = \V^T \x$ are very small for high frequencies.
The smoothest signal is the constant vector $\mathbf{1}$---the first eigenvector $\v_1$ for $\L$ corresponding to the smallest eigenvalue $\lambda_1 = 0$.
Note that $\v_1 = \mathbf{1}$ has one sub-graph of connected nodes of the same sign (strong nodal domain), and higher frequency components $\v_k$ have increasingly more strong nodal domains according to the nodal domain theorem
\cite{biyikoglu05}.

Mathematically, we can write that a signal $\mathbf{x}$ is smooth if its \textit{graph Laplacian regularizer} $\x^T \L \x$ is small \cite{shuman13,pang17}. 
Graph Laplacian regularizer can be expressed as:
\begin{align}
\x^T \L \x  & = \sum_{(i,j) \in \mathcal{E}} w_{i,j} \left( x_i - x_j \right)^2 = \sum_{k} \lambda_k \, \alpha_k^2
\label{eq:smoothness}
\end{align}
Because $\L$ is PSD, $\x^T \L \x$ is lower-bounded by 0.
The graph Laplacian regularizer is related to the \textit{Rayleigh quotient} $R(\x)$, which reaches its minimum at the smallest eigenvalue $\lambda_{\min}$ when $\x = \v_1$,
\begin{equation}
\lambda_{\min} = R(\v_1) = \frac{\v_1^T \L \v_1}{\v_1^T \v_1}
\label{eq:rayleigh}
\end{equation}

We can also interpret the graph Laplacian regularizer as a \textit{signal prior} in a Bayesian formulation; \textit{i.e.}, the probability $Pr(\x)$ of observing a signal $\x$ is:
\begin{equation}
Pr(\x) \propto \exp \left( - \frac{\x^T \L \x}{\sigma^2} \right)
\label{eq:prior}
\end{equation}
where $\sigma$ is a parameter. 
Note that because $\v_1^T \L \v_1 = 0$, the first eigenvector $\v_1$ has the largest probability $Pr(\v_1)$.

\subsection{Graph-Signal Smoothness Prior for Signed Graphs}
\label{subsec:altSmooth}

When considering a more general \textit{signed graph} with positive \textit{and} negative edge weights, conventional graph-signal smoothness priors in the GSP literature can become problematic.
Consider the 3-node line graph in Fig.\;\ref{fig:graphEx1} for $w = 1$ or $-1$.
Using smoothness prior $\x^T \L \x$ in (\ref{eq:smoothness}), we get:
\begin{equation}
\x^T \L \x = -1 (x_1 - x_2)^2 + w (x_2 - x_3)^2
\end{equation}
which promotes a \textit{large} difference between nodes 1 and 2 and a large / small difference between nodes 2 and 3 depending if $w=-1$ or $w=1$.
This prior agrees with notions of inter-node (dis)similarity embedded in edge weights, but direct use of $\x^T \L \x$ can lead to numerical problems. 
For example, $x_1 = \infty$ and $x_2 = -\infty$ would result in $-\infty$, which is a pathological optimal solution for a minimization problem.

Alternatively, one separate GSP approach based on algebraic theory in traditional digital signal processing \cite{sandryhaila13tsp,sandryhaila14,chen14,chen15} interprets the adjacency matrix $\W$ as a shift operator. 
A graph-signal smoothness prior can thus be defined as the difference between the signal $\x$ and its shifted version $\W \x$; specifically, $\left\| \x - \W \x \right\|_p^p$ given a positive integer $p$ was proposed in \cite{chen15}. 
However, when edge weights are negative, this smoothness prior can be insensible. 
For the 3-node graph in Fig.\;\ref{fig:graphEx1}, assuming $p=2$, the smoothness prior when $w=-1$ is:

\begin{align}
\left\| \x - \W \x \right\|_2^2 & =
\left\| (\mathbf{I} - \W) \x \right\|_2^2 =
\left\| \left[ \begin{array}{ccc}
1 & 1 & 0 \\
1 & 1 & 1 \\
0 & 1 & 1
\end{array} \right]
\left[ \begin{array}{c}
x_1 \\
x_2 \\
x_3
\end{array} \right] \right\|_2^2
\nonumber \\
& = (x_1 + x_2)^2 + (x_1 + x_2 + x_3)^2 + (x_2 + x_3)^2 \nonumber
\end{align}
Given two negative edges that signify inter-node dissimilarity, it is not clear why the prior should promote three small sums of signal values.
For example, $\x = (\rho, \rho + 100, \rho)$ for large $\rho > 0$ is clearly a signal with large differences among neighboring pairs, but would compute to a large smoothness prior.

Suppose a \textit{total variation} (TV) approach \cite{rudin92} is taken instead, so that a smoothness prior using $\L$ but based on $l_1$-norm is used instead; \textit{i.e.}, $| \L \x |$. 
Using the same 3-node graph in Fig.\;\ref{fig:graphEx1} with $w=1$, $|\L \x|$ is:

\begin{align}
| \L \x | = \left| 
\left[ \begin{array}{ccc}
-1 & 1 & 0 \\
1 & 0 & -1 \\
0 & -1 & 1 
\end{array} \right]
\left[ \begin{array}{c}
x_1 \\
x_2 \\
x_3
\end{array} \right]
\right|
= \left| \begin{array}{c}
x_2 - x_1 \\
x_1 - x_3 \\
x_3 - x_2
\end{array} \right|  \nonumber
\end{align} 
In other words, the prior tries to minimize the difference between every node pair, even though there is a negative edge between nodes 1 and 2. 
For example, a signal $\x = (\rho, \rho, \rho)$ for some $\rho > 0$ results in $|\L \x| = 0$, but the graph actually demands a large difference between $x_1$ and $x_2$.
Thus this prior is also not sensible.

One final alternative we consider is to adopt a \textit{signed graph Laplacian} definition in \cite{kunegis10}, where $\L^s = \D^s - \W$ and $D^s_{i,i} = \sum_j |w_{i,j}|$. 
Using $\x ^{\top}\L^s \x$ as a smoothness prior, for $w=1$ we get:
\begin{equation}
\x^T \L^s \x = (x_1 + x_2)^2 + (x_2 - x_3)^2
\end{equation}
$w_{1,2} = -1$ means $x_1$ and $x_2$ are expected to take on very different values, but a small $(x_1 + x_2)^2$ only means that $x_1$ and $x_2$ have similar magnitude but opposite signs. 
For example, $x_1 = \rho$ and $x_2 = -\rho$ for very small $\rho > 0$ will also compute to $(x_1 + x_2)^2 = 0$. 
Thus this prior is also not sensible in the general case.

Having demonstrated the shortcomings in alternative smoothness priors in the literature, our goal is to use the original smoothness prior $\x^T \L \x$ (\ref{eq:smoothness}) that agrees with notions of (dis)similarity of edge weights, but perturb Laplacian $\L$ with $\Del$ so that $\L + \Del$ is PSD. 
We discuss this in details in Section\;\ref{sec:algo}.

\subsection{Binary Classifier Graph-Signal Restoration}
\label{subsec:restore}

Given defined Bayesian graph-signal smoothness prior (\ref{eq:prior}), we can now formally define a restoration problem for a binary classifier via a MAP formulation. 
First, to model noise in binary labels, we adopt a uniform noise model \cite{brew10}, where the probability of observing $y_i=x_i$, $1 \leq i \leq K$, is $1-p$, and $p$ otherwise; \textit{i.e.},
\begin{equation}
Pr(y_i | x_i) = \left\{ \begin{array}{ll}
1-p & \mbox{if} ~~ y_i = x_i \\
p & \mbox{o.w.}
\end{array} \right.
\label{eq:noise}
\end{equation}
This noise model is motivated by the following observation in social media analysis, 
when labels are often assigned manually by non-experts via \textit{crowd-sourcing} \cite{brew10}---\textit{i.e.} employ many non-experts online to assign labels to a subset of data at a very low cost.
Because non-experts can be unreliable (\textit{e.g.}, a non-expert is not competent in a label assignment task but pretends to be, or he simply assigns label randomly to minimize mental effort), observations $\y$ may result in label errors or noise that are uniform and independent. 

The probability of observing a noise-corrupted $\mathbf{y}$, $\y \in \{-1, 1\}^K$, given ground truth $\mathbf{x}$, $\x \in \{-1, 1\}^N$, where $K<N$, is:
\begin{align}
Pr(\y | \x) & = p^k (1-p)^{K-k} \nonumber \\
k & = \| \mathbf{y} - \H \x \|_0
\label{eq:likelihood}
\end{align}
where $\H \in \{0, 1\}^{K \times N}$ is a matrix that picks $K$ observations from $N$ total samples.  
(\ref{eq:likelihood}) serves as the likelihood term given label noise model in (\ref{eq:noise}).
The negative log of this likelihood $Pr(\y | \x)$ can be rewritten as:
\begin{align}
- \log Pr(\mathbf{y} | \mathbf{x}) =
k \underbrace{\left( \log(1-p) - \log(p) \right)}_{\gamma} - K \log(1-p)
\end{align}
Because the second term is a constant for fixed $K$ and $p$, we can ignore it during minimization.

Given prior (\ref{eq:smoothness}) and likelihood (\ref{eq:likelihood}), we can formulate a MAP problem as follows:
\begin{equation}
\min_{\mathbf{x}}
\| \y - \H \x \|_0 + \mu \; \mathbf{x}^T \left( \mathbf{L} + \Del \right) \mathbf{x}
\label{eq:obj1}
\end{equation}
where $\mu$ is a parameter that trades off the importance between the likelihood term and the signal prior, and $\Del$ is the perturbation matrix to be discussed in Section\;\ref{sec:algo}.
We discuss how (\ref{eq:obj1}) can be solved efficiently in Section\;\ref{sec:gsmooth}.


\section{Graph Construction}
\label{sec:construct}
\subsection{Construct Graph with Negative Edges}
\label{subsec:construct}


\begin{figure}

\begin{minipage}[b]{.48\linewidth}
 \centering
 \centerline{\includegraphics[width=3.6cm]{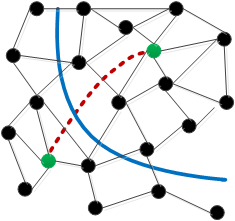}}
 \vspace{0.10cm}
 \centerline{a) centroid-based graph}\medskip
\end{minipage}
\hfill
\begin{minipage}[b]{.48\linewidth}
 \centering
 \centerline{\includegraphics[width=3.7cm]{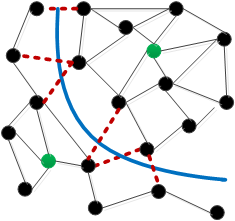}}
 \vspace{-0.10cm}
 \centerline{b) boundary-based graph}\medskip
\end{minipage}

\vspace{-0.1in}
\caption{Two constructions of similarity graphs with negative edges: a) connect cluster centroids (in red), and b) connect boundary nodes of two clusters. A blue line denotes the border between two clusters in each graph.}
\label{fig:negGraph}

\end{figure}

In a semi-supervised learning problem, often we are given only a \textit{feature vector} for each sample in a high-dimensional feature space, with (possibly noisy) labels assigned to a small sample subset.
To compute a graph-based binary classifier, we must first construct an appropriate graph where edges reflect inter-node (dis)similarity relationships based on features.
We discuss our graph construction strategy here.
\textit{To the best of our knowledge, the construction of similarity graphs with both positive and negative edges from feature vectors for classification has not been studied in the graph-based classifier literature.}

We first construct a graph $\cG$ with nodes $\cV$ representing $N$ samples.
For each sample $i$ we assume that there exists a corresponding feature vector $\h_i$ of dimension $Q$.
Then we can assign \textit{positive} edge weight $w_{i,j}$ using a Gaussian kernel:
\begin{equation}
w_{i,j} = \exp \left(- \frac{(\h_i - \h_j)^T \Xii (\h_i - \h_j)}{\sigma_h^2} \right)
\label{eq:edgeWeight}
\end{equation}
where $\sigma_h$ is a parameter.
$\Xii$ is a $Q \times Q$ diagonal matrix, where $\Xi_{i,i}$ is a feature weight for the $i$-th feature.
We assign positive edges with weights $w_{i,j}$ to connect node $i$ to its $\omega$'s nearest neighbors\footnote{If this relationship is not symmetric, \textit{i.e.}, if $i$ is one of $\omega$ closest neighbors to $j$ but $j$ is not one of $\omega$ closest neighbors to $i$, then we keep edge $(i,j)$ of weight $w_{i,j}$ anyway. Thus each node has $\geq \omega$ neighbors.} $j$.

This positive weight assignment is similar to those in previous works on spectral clustering \cite{shi00} and graph-based classifier learning \cite{gadde14,mao16}, where a closer distance in the $Q$-dimensional feature space leads to a larger edge weight.
To improve clustering / classification performance, feature parameter $\Xi_{i,i}$ is set larger if the $i$-th feature is more discriminate.
For optimization of feature parameters $\Xi_{i,i}$---which is not the focus of this paper---see \cite{huang05}.

We propose two methods to insert negative edges into an initial graph with only positive edges.
The first results in a graph $\cG^1$ that is robust to label noise but not precise in designating inter-node dissimilarity relationships, and the second results in a graph $\cG^2$ that is precise in designating dissimilarity relationships but not robust to noisy labels.

In the first \textit{centroid-based} method, we divide the samples into two similar clusters based on observed labels (or estimated labels from previous iteration), then connect the two respective \textit{centroids} with a negative edge, as illustrated in Fig.\;\ref{fig:negGraph}(a).
The idea is that even if some sample labels are corrupted by noise, given that the two clusters are sufficiently different, then at least the \textit{centers} (centroids) of the two clusters are expected to have opposing labels.
However, ideally the \textit{boundaries} of the clusters should define the label crossover points.
Thus $\cG^1$ is robust but not precise.

In the second \textit{boundary-based} method, we connect the boundary samples of the two clusters with negative edges, as illustrated in Fig.\;\ref{fig:negGraph}(b).
This construction leads to enhancement of the cluster boundaries during filtering and thus improves classification performance.
However, given that the labels are noise-corrupted, the exact locations of the boundaries are initially uncertain, and hence $\cG^2$ is not precise.

We thus propose to combine the two graphs as follows and iterate.
For each graph, we construct a graph Laplacian $\L^i$ and compute a suitable perturbation matrix $\Del^i$ (to be discussed in Sections \ref{sec:algo} and \ref{sec:fast}).
We then combine them as a convex combination:
\begin{align}
\L^* = \beta \left( \L^1 + \Del^1 \right) + (1-\beta) \left( \L^2 + \Del^2 \right)
\end{align}
where $0 \leq \beta \leq 1$ is a parameter that changes from 1 to 0 as we iterate.
Thus $\L^*$ will be robust early in the iterations, and precise late in the iterations.

\subsection{Example of Graph with Negative Edges}
\label{subsec:graphexample}
\begin{figure}

\begin{minipage}[b]{.52\linewidth}
 \centering
 \centerline{\includegraphics[width=4.7cm]{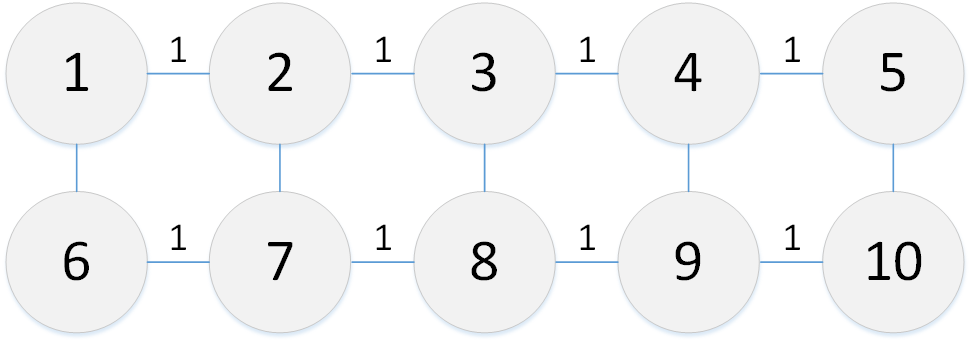}}
 \vspace{0.2cm}
 \centerline{a) 10-node graph}\medskip
\end{minipage}
\hfill
\begin{minipage}[b]{.45\linewidth}
 \centering
 \centerline{\includegraphics[width=4.0cm]{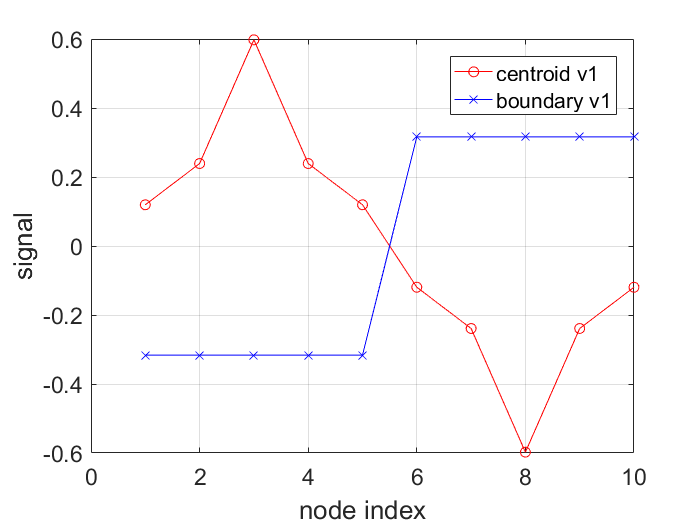}}
 \vspace{-0.15cm}
 \centerline{b) first eigenvectors}\medskip
\end{minipage}

\vspace{-0.1in}
\caption{Example of a 10-node graph. Nodes $1$ to $5$ ($6$ to $10$) belong to one class and are connected by edges of weight $1$. For centroid-based graph, nodes between the two classes are connected by edges of weight $0.1$, except $(3,8)$ which is connected by an edge of weight $-1$. For boundary-based graph, nodes between the two classes are connected by edges of weight $-1$.}
\label{fig:graph10node}

\end{figure}

As illustration, we consider a simple example in Fig.\;\ref{fig:graph10node}(a): a 10-node graph where nodes $1$ through $5$ are similar and are connected by edges of weight $1$, and nodes $6$ to $10$ are similar.
For the centroid-based graph, nodes between the two classes are connected by edges of weight $0.1$, except the two respective centroids $(3,8)$ that are connected by an edge of weight $-1$.
Graph Laplacian $\L$ for this graph has smallest eigenvalue $-0.8$, and the corresponding eigenvectors $\v_1$ is shown in Fig.\;\ref{fig:graph10node}(b).
We see that the maximum and minimum of $\v_1$ are located at the endpoints $(3,8)$ of the lone negative edge, and thus during signal restoration, the prior will promote opposite label assignments for samples $3$ and $8$, which agrees with the dissimilarity notion of negative edges.
\textit{More generally, low graph frequency components $\v_i$ of an indefinite graph Laplacian $\L$ are useful in restoring signal $\x$, leveraging inter-node dissimilarity information embedded in negative edges.} This point is also argued in \cite{knyazev17arxiv} for spectral clustering.

For the boundary-based graph, boundary nodes between the two classes are all connected by edges of weight $-1$. Graph Laplacian $\L$ for this graph has smallest eigenvalue $-2$, and the corresponding eigenvectors $\v_1$ is also shown in Fig.\;\ref{fig:graph10node}(b).
We see that each pair of boundary nodes across two clusters have the same opposite values, and thus during restoration, the prior will promote opposite label assignments for all negative-edge-connected pairs.

\section{Finding a Perturbation Matrix}
\label{sec:algo}
We now address the problem of identifying a perturbation matrix $\Del$ such that $\L + \Del$ is PSD. 
To impart intuition on the effects of $\Del$ on the eigenvalues of $\L + \Del$, consider first the 
\textit{Weyl's inequality}~\cite{weyl12}. 
Let a Hermitian matrix $\L \in \mathbb{R}^{N \times N}$ have spectral decomposition $\L = \V \Lamb \V^T$ as described in (\ref{eq:spectrum}), with eigenvalues $\lambda_k$ along the diagonal of diagonal matrix $\Lamb$. 
Let $\Del$ be a Hermitian matrix with eigenvalues $\gamma_1 \leq \ldots \leq \gamma_N$. 
Weyl's inequality states that $\L + \Del$ has eigenvalues $\nu_1 \leq \ldots \leq \nu_N$, such that:
\begin{equation}
\lambda_i + \gamma_1 \leq \nu_i \leq \lambda_i + \gamma_N
\label{eq:weyl}
\end{equation}
In words, (\ref{eq:weyl}) states that the $i$-th eigenvalue $\nu_i$ of $\L + \Del$ is the $i$-th eigenvalue $\lambda_i$ of $\L$ shifted by an amount in the range $[\gamma_1, \gamma_N]$. 
Assume that $\lambda_1 < 0$, hence $\L$ is indefinite. 
The Weyl's inequality then implies that for $\L + \Del$ to be PSD:
i) a necessary (but not sufficient) condition is $\gamma_N \geq -\lambda_1$;
ii) a sufficient (but not necessary) condition is $\gamma_1 \geq -\lambda_1$.

Obviously, there exists an infinite number of feasible solutions $\Del$.
Thus a well chosen criteria must be used to differentiate them.

\subsection{Matrix Perturbation: Minimum-Norm Criteria}

One reasonable choice is the \textit{minimum-norm criteria}, \textit{i.e.}, find $\Del$ with the smallest norm such that $\L + \Del$ is PSD:
\begin{equation}
\min_{\Del} \left\| \Del \right\|
~~~ \mbox{s.t.} ~~ \x^T \left( \L + \Del \right) \x \geq 0, ~ \forall \x
\label{eq:minNorm}
\end{equation}
where $\| . \|$ is a \textit{unitarily invariant norm} on $\mathbb{R}^{N \times N}$; \textit{i.e.} $\| \U \Del \V \| = \| \Del \|$ for all orthogonal $\mathbf{U}$ and $\mathbf{V}$. 

It turns out that the solution to (\ref{eq:minNorm}) is a special case of 
Theorem 5.1 in \cite{higham98}, which we rephrase as follows.
Assume that $\L$ has exactly $p$ negative eigenvalues.
Theorem 5.1 in \cite{higham98} states that the optimal perturbation matrix $\Del$ with minimum norm $\| \Del \|$, such that $\L + \Del$ is PSD, is: 
\begin{equation}
\Del = \V ~ \mathrm{diag}(\boldsymbol{\tau}) ~ \V^T 
\label{eq:minNormDel}
\end{equation}
where $\boldsymbol{\tau} = [\tau_1, \ldots, \tau_n]$:
\begin{equation}
\tau_i = \left\{ \begin{array}{ll}
- \lambda_i & \mbox{if} ~~ 1 \leq i \leq p \\
0 & \mbox{o.w.}
\end{array} \right.
\end{equation}

See \cite{higham98} for a complete proof. 
We only make a few important observations.
First, it is clear that $\L + \Del$ is PSD:
\begin{align}
\L + \Del & = \V 
\mathrm{diag}(\lambda_1 - \lambda_1, \ldots, \lambda_p - \lambda_p, \lambda_{p+1}, \ldots, \lambda_n)
\V^T \nonumber \\
& = \V \mathrm{diag}(0, \ldots, 0, \lambda_{p+1}, \ldots, \lambda_n) \V^T \nonumber
\end{align}
Since all the negative eigenvalues of $\L$ have been eliminated, $\L + \Del$ is PSD.

Second, due to the definition (\ref{eq:minNormDel}) of $\Del$, $\L + \Del$ can be spectrally decomposed using the \textit{same} eigenvectors $\V$ as original $\L$. 
As discussed previously and also argued in \cite{knyazev17arxiv} for spectral clustering, maintaining the same eigen-space in the perturbed matrix $\L + \Del$---especially the low graph frequency components---is desirable.

Third, by eliminating all negative eigenvalues of $\L$ to $0$, the first $p+1$ eigenvectors $\v_1, \ldots, \v_p$ will all evaluate to $0$ in the quadratic regularizer: 
\begin{equation}
\v_i^T \left( \L + \Del \right) \v_i = 0, ~~ 1 \leq i \leq p + 1
\end{equation}
$p+1$ because $\L$ contains the DC component $\mathbf{1}$ that the regularizer also evaluates to 0.
Hence the regularizer expresses no preference among the first $p+1$ eigenvectors.
This is problematic during graph-signal restoration. 
This means that though the graph structure $\mathcal{G}$ has a notion of frequencies and the original (numerically unstable) smoothness prior prefers low frequencies, the augmented regularizer does not differentiate and maps the lowest $p+1$ frequencies all to zero. 
This is clearly sub-optimal.

\subsection{Matrix Perturbation: Eigen-structure Preservation}

The main problem with the minimum-norm criteria is that the differentiation among different frequency components (eigenvectors) is removed by setting all negative eigenvalues of $\L$ to $0$.
Thus, it is desirable to perturb $\L$ in a way that the frequency components \textit{and} their frequency preferences (\textit{i.e.}, low frequencies are still preferred over high frequencies) are preserved, while the overall perturbation is minimized.

Suppose first that we want to preserve the entire eigen-structure of $\L$: eigenvectors of $\L$ and spacings between neighboring eigenvalues of $\L$ during perturbation.  
In other words, for perturbed $\L + \Del$, we require
\begin{equation}
\lambda_{i+1} - \lambda_i = \nu_{i+1} - \nu_i.
\end{equation}
One method of achieving this, leveraging on the Weyl's inequality~\cite{weyl12}, is to select perturbation matrix $\Del = \eta \, \I$, for some $\eta > 0$, so that $\Del$ has eigenvalue $\eta$ with multiplicity $N$.
Clearly eigenvectors of $\L$ are preserved in $\L + \Del = \V (\L + \eta \I) \V^T$, and eigenvalue spacings are also preserved: 
$\nu_i = \lambda_i + \eta$ and $\nu_{i+1} - \nu_i = \lambda_{i+1} - \lambda_i$.  

Given $\Del = \eta \, \I$, one interpretation of the smoothness prior $\x^T (\L + \Del) \x$ is that it is a weighted sum of signal variations and signal energies:
\begin{align}
\x^T (\L + \Del) \x & = \x^T \L \x + \eta\, \x^T \I \x \nonumber \\
& = \sum_{i,j} w_{i,j} (x_i - x_j)^2 + \eta \sum_i x_i^2
\end{align}
In other words, to make smoothness prior $\x^T \L \x$ numerically stable, we consider in addition a weighted signal energy term to avoid pathological solutions like $-\infty \v_1$ for $\lambda_1 < 0$. 

To find the perturbation matrix $\Del = \eta \, \I$ with the minimum norm $\| \Del \|$ such that $\L + \Del$ is PSD, we only need to identify the smallest eigenvalue $\lambda_{\min} = \lambda_1 < 0$ of $\L$ and set $\eta = - \lambda_{\min}$. 
Computing $\lambda_{\min}$ directly can be computationally expensive, however; we discuss faster methods to compute lower bounds for $\lambda_{\min}$ next.

\subsection{A Simple Lower Bound for $\lambda_{\min}$}

We can compute a lower bound for $\lambda_{\min}$ simply as follows.
Denote by $\L^+$ and $\L^-$ the graph Laplacian matrices corresponding to edges with positive and negative weights in graph $\cG$ respectively; clearly $\L = \L^+ + \L^-$.
The Rayleigh quotient for $\L$ can be expanded as:
\begin{equation}
\frac{\x^T \L \x}{\x^T \x} = 
\frac{\x^T \left( \L^+ + \L^- \right) \x}{\x^T \x}
\end{equation}

Because $\L^+$ containing only positive edges is PSD, the first term in the numerator $\x^T \mathbf{L}^+ \x$ is lower-bounded by $0$.
For the second term, we can first define $\cL^- = - \L^-$, which is PSD, and write:
\begin{align}
\frac{\x^T \L^- \x}{\x^T \x} &= 
- \left( \frac{\x^T \cL^- \x}{\x^T \x} \right) \geq - \lambda^-_{\max}
\end{align}
where $\lambda^-_{\max}$ is the largest eigenvalue of $\cL^-$. 
Since $-\lambda^-_{\max}$ is also the lower bound of the Rayleigh quotient for $\L$, it is also the lower bound for the smallest eigenvalue $\lambda_{\min}$ of $\L$:
\begin{equation}
-\lambda^-_{\max} \leq \lambda_{\min}
\label{eq:lbound}
\end{equation}
Thus a perturbation matrix $\Del = \lambda^-_{\max} \mathbf{I}$ would result in $\L + \Del$ that is PSD.
$\lambda^-_{\max}$ for matrix $\cL^-$ can be computed using the \textit{power iteration method}, which has complexity $O(N)$ for a sparse graph per iteration \cite{golub12}.
However, convergence speed depends on the distance between $\lambda^-_{\max}$ and the next largest eigenvalue $\lambda_{N-1}^-$ in $\cL^-$; \textit{i.e.}, smaller $|\lambda_{\max}^- - \lambda_{N-1}^-|$ means a slower convergence rate.  
Further, this lower bound (\ref{eq:lbound}) is often loose in practice.
We next discuss a faster and more robust computation of a lower bound for $\lambda_{\min}$.

\section{Fast Computation}
\label{sec:fast}

Our goal is to obtain a lower bound $\lambda_{\min}^{\#}$ for $\lambda_{\min}$ robustly and efficiently.  
Having obtain $\lambda_{\min}^{\#}$, we can add perturbation matrix $\Del^{\#} = -\lambda_{\min}^{\#} \I$ to $\L$, so that the resulting $\L + \Del^{\#}$ is PSD.
State-of-the-art numerical linear algebra methods include Lanczos method and its variants \cite{golub12}; given prior knowledge about the interval in which a desired eigenvalue lives, one can compute it using shift-and-invert Lanczos methods, for example. 
However, these methods require prior knowledge about the existence of eigenvalues at different intervals. 
In our proposal, no such prior knowledge is required.

\subsection{Matrix Inertia}

We first define matrix inertia.
The \textit{inertia} $\mathrm{In}(\A)$ of a matrix $\A$
is a set of three numbers counting the positive, negative, and zero eigenvalues in $\A$:
\begin{equation}
\mathrm{In}(\A) = \left( i^{+}(\A),~ i^{-}(\A),~ i^{0}(\A) \right)
\end{equation}
where $i^{+}(\A)$, $i^{-}(\A)$ and $i^{0}(\A)$ denote respectively the number of positive, negative and zero eigenvalues in matrix $\A$.
Inertia is an intrinsic property of the matrix; 
according to \textit{Sylvester's Law of Inertia}\footnote{https://en.wikipedia.org/wiki/Sylvester\%27s\_law\_of\_inertia}, the inertia of a matrix is invariant to any congruent transform, \textit{i.e.},
\begin{equation}
\mathrm{In}(\A) = \mathrm{In}(\P^T \A \P)
\end{equation}
where $\P$ is an invertible matrix.

\subsection{Graph Partition}

To reduce complexity, we can divide the node set $\cN$ into two subsets $\cN_1$ and $\cN_2$,
so that intensive computation is performed in the node subsets separately.
Note that partitioning a graph into two node sets to reduce complexity is also done in \textit{Kron reduction} \cite{dorfler13}.
However, \cite{dorfler13} considers only PSD $\L$ (possibly with self-loops), while we consider indefinite $\L$ that requires perturbation $\Del$ to make $\L + \Del$ PSD.

Given the two sets $\cN_1$ and $\cN_2$, we can write the graph Laplacian $\L$ in blocks:
\begin{equation}
\L = \left[ \begin{array}{cc}
\L_{1,1} & \L_{1,2} \\
\L_{1,2}^T & \L_{2,2}
\end{array}
\right]
\label{eq:glPartition}
\end{equation}
where $\L_{1,1}$ and $\L_{2,2}$ are sub-matrices of respective dimension $|\cN_1| \times |\cN_1|$ and $|\cN_2| \times |\cN_2|$ corresponding to node sets $\cN_1$ and $\cN_2$, and $\L_{1,2}$ is a $|\cN_1| \times |\cN_2|$ sub-matrix corresponding to cross-connections between $\cN_1$ and $\cN_2$.

We can now relate the inertia of $\L$ with its sub-matrices using the \textit{Haynsworth Inertia Additivity} formula \cite{haynsworth68}:
\begin{equation}
\mathrm{In}(\L) = \mathrm{In}(\L_{1,1}) + \mathrm{In}(\L / \L_{1,1})
\label{eq:haynsworth}
\end{equation}
where $\L / \L_{1,1}$ is the \textit{Schur Complement}\footnote{https://en.wikipedia.org/wiki/Schur\_complement} (SC) of block $\L_{1,1}$ of matrix $\L$ in (\ref{eq:glPartition}), which is defined as
\begin{equation}
\L / \L_{1,1} = \L_{2,2} - \L_{1,2}^T \L_{1,1}^{-1} \L_{1,2}
\label{eq:SC}
\end{equation}
Thus, if we can ensure that $\L_{1,1}$ \textit{and} its SC do not contain negative eigenvalues, then $\L$ will also have no negative eigenvalues and is PSD.
We develop an efficient algorithm based on this idea next.

\subsection{Eigenvalue Lower Bound Algorithm}
\label{subsec:LowerBound}

\begin{figure}

\begin{minipage}[b]{.95\linewidth}
 \centering
 \centerline{\includegraphics[width=6.5cm]{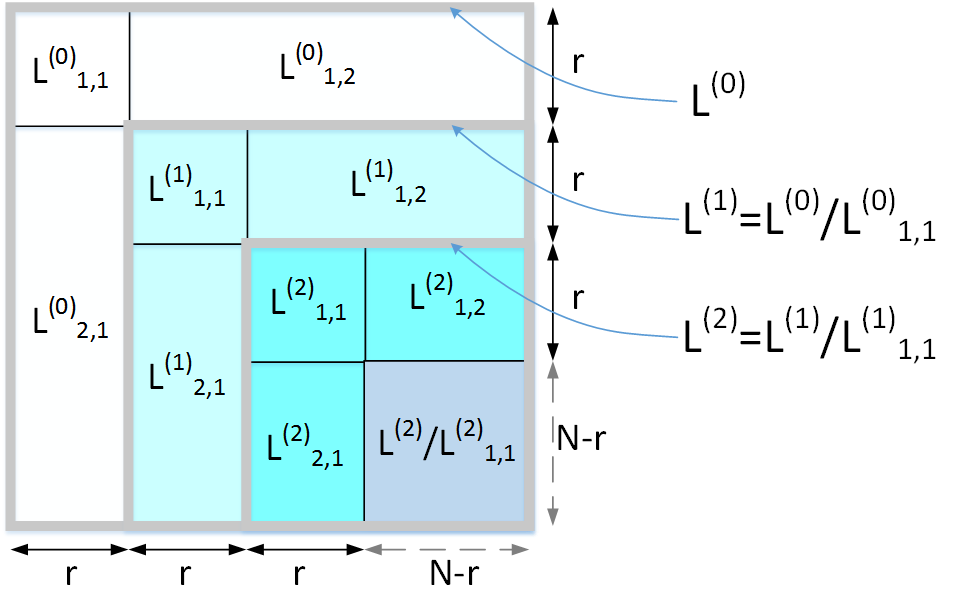}}
\end{minipage}

\vspace{-0.15in}
\caption{Example of recursively partitioning Schur complement $\L^t / \L^{t}_{1,1}$ into two set of nodes: $\L^1 = \L^0 / \L^0_{1,1}$ into $\L^1_{1,1}$ and $\L^1 / \L^1_{1,1}$, then $\L^2 = \L^1 / \L^1_{1,1}$ into $\L^2_{1,1}$ and $\L^2 / \L^2_{1,1}$.}
\label{fig:fastAlgo}

\end{figure}

We propose the following recursive algorithm to find a lower bound $\lambda_{\min}^{\#}$ for $\L$.
See Fig.\;\ref{fig:fastAlgo} for an illustration.
We initialize $t:=0$ and $\L^{0} := \L$.
We define a recursive algorithm $EvalBound(\L^t, t)$ that returns a lower bound $\lambda_{\min}^t$ for eigenvalues in $\L^t$.
It has two steps as described below.

\vspace{0.1in}
\noindent\textbf{Step 1}: ~
We first partition node set $\cN^t$ in $\L^t$ into two subsets $\cN^t_1$ and $\cN^t_2$, where $|\cN^t_1| = r$. $r$ is a pre-defined parameter to control computation complexity.
$\cN^t_1$ can be chosen by first randomly selecting a node in $\cN^t$, then perform \textit{breadth-first search} (BFS) \cite{clr} until $r$ nodes are discovered.
We eigen-decompose $\L^{t}_{1,1}$ to find its smallest eigenvalue $\lambda^{t}_1$.
We define the \textit{augmented eigenvalue} $\kappa_{\min}^{t}$ as:
\begin{equation}
\kappa_{\min}^{t} = \left\{ \begin{array}{ll}
\lambda^{t}_1 - \epsilon & \mbox{if} ~~~ \lambda^{t}_1 \leq 0 \\
0 & \mbox{o.w.}
\end{array} \right.
\label{eq:kappa}
\end{equation}
where $\epsilon > 0$ is a small parameter.
We perturb matrix $\L^t$ using computed $\kappa_{\min}^t$, \textit{i.e.} $\cL^{t} = \L^{t} - \kappa_{\min}^{t} \I$.
It is clear that $\cL^{t}_{1,1}$ is positive definite (PD) and thus invertible.

\vspace{0.1in}
\noindent\textbf{Step 2}: ~
We ensure SC of $\cL^{t}_{1,1}$ of $\cL^t$ is PSD. By definition, the SC is:
\begin{equation}
\cL^{t}/\cL^{t}_{1,1} = \cL_{2,2}^{t}
- (\L_{1,2}^{t})^T (\cL_{1,1}^{t})^{-1} \L_{1,2}^{t}
\label{eq:sc}
\end{equation}
$\cL^{t}/\cL^{t}_{1,1}$ can be interpreted as a $|\cN_2^{t}| \times |\cN_2^{t}|$ graph Laplacian matrix for nodes $\cN_2^{t}$.
If $|\cN_2^{t}| \leq r$, then we eigen-decompose $\cL^{t}/\cL^{t}_{1,1}$ and find its smallest eigenvalue $\lambda_2^{t}$.
We compute $\lambda_{\min}^t := \kappa_{\min}^t + \min \left(\lambda_2^{t}, 0 \right)$.
We exit the algorithm with $\lambda_{\min}^t$ as solution.

If $|\cN_2^{t}| > r$, we set $\L^{t+1} := \cL^{t}/\cL^{t}_{1,1}$ and recursively call $\eta_{\min}^t := EvalBound(\L^{t+1}, t+1)$.
Upon return, we compute $\lambda_{\min}^t := \kappa_{\min}^t + \eta_{\min}^t$ and exit the algorithm with $\lambda_{\min}^t$ as solution.

\subsection{Proof of Algorithm Correctness}

We now prove that $EvalBound(\L, 0)$ returns a lower bound for the true minimum eigenvalue $\lambda_{\min}$ of $\L$.
Specifically, we prove by induction the following recursion invariant:
\textit{At each recursive call $t$, given $\L^t$, the computed $\lambda_{\min}^t$ is a lower bound for eigenvalues of $\L^t$.}

We first examine the base case. At a leaf recursive call $\tau$, in Step 1, $\L^{\tau}$ is perturbed using computed $\kappa_{\min}^{\tau}$ such that $\cL_{1,1}^{\tau} = \L_{1,1}^{\tau} - \kappa_{\min}^{\tau} \I$ is PD.
In Step 2, if computed $\lambda_{2}^{\tau} \geq 0$, then SC $\cL^{\tau} / \cL_{1,1}^{\tau}$ is PSD.
Since $\cL_{1,1}^{\tau}$ and its SC $\cL^{\tau} / \cL_{1,1}^{\tau}$ are both PSD, by (\ref{eq:haynsworth}) $\cL^{\tau}$ is also PSD.
Hence $\lambda_{\min}^{\tau} = \kappa_{\min}^{\tau}$ is a lower-bound for matrix $\L^{\tau}$.

If $\lambda_{2}^{\tau} < 0$, then perturbed SC, $\cL^{\tau} / \cL_{1,1}^{\tau} - \lambda_{2}^{\tau} \I$, is PSD.
It turns out that if we perturb $\cL^{\tau}$ again using $\lambda_2^{\tau}$, \textit{i.e.}, $\cL'^{\tau} = \cL^{\tau} - \lambda_2^{\tau} \I$, then $\cL'^{\tau}$ is PSD.
This is because $\cL_{1,1}'^{\tau} = \cL_{1,1}^{\tau} - \lambda_2^{\tau} \I$ is PD, and its SC $\cL'^{\tau} / \cL_{1,1}'^{\tau}$ is PSD by the following lemma:

\begin{lemma}
If $\L_{1,1}$ is PD and $\L/\L_{1,1} + \delta \I$ is PSD for $\delta > 0$, then SC $\L'/\L'_{1,1}$, where $\L' = \L + \delta \I$, is also PSD.
\label{lemma:2}
\end{lemma}
See Appendix for a full proof.
In this case $\lambda_{\min}^{\tau} = \kappa_{\min}^{\tau} + \lambda_2^{\tau}$, hence $\lambda_{\min}^{\tau}$ is a lower bound for $\L^{\tau}$.

Consider now the inductive case, where at iteration $t$ we assume that, $\eta_{\min}^t := EvalBound(\L^{t+1}, t+1)$ is a lower bound for $\L^{t+1} = \cL^t / \cL^t_{1,1}$.
From Step 1, we know that $\L^t$ is perturbed using $\kappa_{\min}^t$ so that $\cL_{1,1}^t = \L_{1,1}^t - \kappa^t_{\min} \I$ is PD.
By assumption, we know that $\cL^t / \cL_{1,1}^t$ can be perturbed using $\eta_{\min}^t$ such that $\cL^t / \cL_{1,1}^t - \eta_{\min}^t \I$ is PSD.
By Lemma \ref{lemma:2}, we know that $\cL^t - \eta_{\min}^{t} \I$ is also PSD.
Thus $\lambda_{\min}^t := \kappa_{\min}^t + \eta_{\min}^t$ is a lower bound for $\L^t$.

Since both the base case and the inductive case are proven, the recursion invariant is also proven, and $EvalBound(\L, 0)$ returns a lower bound for $\lambda_{\min}$ of $\L$. $\Box$

\subsection{Computation Complexity}

\label{subsec:complexity}
We can estimate the computation cost of our algorithm as follows.
For each recursive call, the cost of eigen-decomposing a $r \times r$ matrix is $O(r^3)$ operations.
The number of recursive calls is $O(N / r)$.
Thus the complexity of step 1 of our algorithm  is $O((N/r)r^3) = O(N r^2)$.

The cost of computing SC in (\ref{eq:sc}) can be bounded as follows. 
In the extreme case when $r=1$, the off-diagonal blocks $\L^{t}_{1,2}$ are vectors, and thus computing (\ref{eq:sc}) throughout the algorithm means that each off-diagonal entry in $\L$ is accessed exactly once, resulting in $O(N^2)$. 
However, if we assume the original Laplacian $\L$ is sparse, then only $O(N)$ entries in $\L$ are non-zero, reducing the complexity to $O(N)$.
When $r > 1$, each entry in $r \times r$ matrix $(\mathcal{L}^t_{1,1})^{-1}$ will access an entry in block $\L^{t}_{1,2}$, resulting in complexity $O(N r^2)$. 
Thus the overall complexity is $O(N r^2)$. 
Compared to the complexity $O(N^3)$ of eigen-decomposition of the larger matrix $\L$, this represents a non-trivial computation saving.





\section{Classifier Learning with Noisy Labels}
\label{sec:gsmooth}
Having discussed a fast method to compute $\Del$ such that $\L + \Del$ is PSD, we now discuss how to optimize (\ref{eq:obj1}).

But first, beyond the graph-signal smoothness prior we defined in Section\;\ref{subsec:smooth}, we in addition define a \textit{generalized smoothness prior} in Section \ref{subsec:gsmooth}, which promotes ambiguity in the classifier signal.
We provide a novel interpretation of generalized smoothness on graph by viewing a graph-signal as voltages on an electrical circuit in Section \ref{subsec:circuit}.
Finally, after formulating the problem with both prior terms in Section\;\ref{subsec:obj2}, we propose an algorithm to solve it efficiently in Section\;\ref{subsec:IRLS}.

\subsection{Generalized Smoothness}
\label{subsec:gsmooth}


\subsubsection{Positive Edge Weights for Generalized Smoothness}

Like TGV for images \cite{bredies15}, one can also define a higher-order notion of smoothness for graph-signals using \textit{positive} edge weights \cite{mao16}.
Specifically, positive edge graph Laplacian $\L^+$ is related to the second derivative of continuous functions \cite{shuman13}, and so $\L^+ \mathbf{x}$ computes the second-order difference on graph-signal $\mathbf{x}$. As an example, the 3-node line graph in Fig.\;\ref{fig:graphEx1} with $w=1$ has the following $\L^+$:
\begin{equation}
\L^+ = \left[
\begin{array}{ccc}
1 & -1 & 0 \\
-1 & 2 & -1 \\
0 & -1 & 1
\end{array}
\right]
\end{equation}
Using the second row $\L^+_{2,:}$ of $\mathbf{L}^+$, we can compute the second-order difference at node $x_2$:
\begin{equation}
\L^+_{2,:} \mathbf{x} = - x_1 + 2 x_2 - x_3
\label{eq:ex1a}
\end{equation}

On the other hand, the definition of second derivative of a function $f(x)$ is:
\begin{equation}
f''(x) = \lim_{h \rightarrow 0} \frac{f(x+h) - 2 f(x) + f(x-h)}{h^2}
\label{eq:ex1b}
\end{equation}
We see that (\ref{eq:ex1a}) and (\ref{eq:ex1b}) are computing the same quantity (with a sign change) in the limit.

Hence if $|\L^+ \mathbf{x}|$ is small, then the second-order difference of $\mathbf{x}$ is small, or the first-order difference of $\mathbf{x}$ is smooth or changing slowly.
In other words, the \textit{gradient} of the signal is smooth with respect to the graph.
We express this notion by stating that the square of the $l_2$-norm of $\L^+ \mathbf{x}$ is small:
\begin{equation}
\| \L^+ \mathbf{x} \|_2^{2} = \mathbf{x}^T (\L^+)^T \L^+ \mathbf{x}
= \mathbf{x}^T (\L^+)^2 \mathbf{x} = \sum_i (\lambda_i^+)^2 \alpha_i^2
\label{eq:GTV}
\end{equation}
where (\ref{eq:GTV}) is true since $\L^+$ is symmetric by definition\footnote{Note that powers of the graph Laplacian $\L$ have been used previously to achieve signal smoothness within a local neighborhood \cite{milanfar13,mao16tmm}.}.

\subsubsection{Negative Edges for Generalized Smoothness}

We now argue that using an indefinite graph Laplacian $\L$ containing negative edges to define generalized smoothness $\x^T \L^2 \x$ is problematic. 
One reason is that while the frequency components are preserved,
\begin{equation}
\L^2 = \V \Lamb \V^T \V \Lamb \V^T = \V \Lamb^2 \V^T
\end{equation}
frequency preferences are reordered in $\L^2$; \textit{e.g.}, an eigenvector $\v_i$ corresponding to a negative eigenvalue $\lambda_i < 0$ in $\L$ now corresponds to eigenvalue $\lambda_i^2 > 0$ in $\L^2$. Thus $\v_i$ and a possible eigenvector $\v_j$ that corresponds to a positive eigenvalue $\lambda_j > 0$ with magnitude smaller than $\lambda_i$ have switched order when ordered from smallest corresponding eigenvalues to largest in $\L^2$.
It is hard to explain how this reordering of frequency components according to magnitude $\lambda_i^2$ is beneficial for signal restoration.

To illustrate the potential problem of negative edges in generalized smoothness in the nodal domain, consider again the three-node line graph in Fig.\;\ref{fig:graphEx1}, where $w = 1$.
The corresponding second row of the graph Laplacian $\L$ is:
\begin{equation}
\L_{2,:} = \left[ \begin{array}{ccc}
1 & 0 & -1
\end{array} \right]
\end{equation}
This means that when we compute the generalized smoothness $|\L \x|$ at $x_2$, we get $|\L_{2,:} \x| = |x_1 - x_3|$; \textit{i.e.,} the generalized smoothness at $x_2$ does not actually depend on the value of $x_2$, which is not sensible.
We provide an alternative explanation why negative edge weights should not be used next.

\subsection{Circuit Interpretation of Generalized Smoothness}
\label{subsec:circuit}

It has been shown that by interpreting an undirected weighted graph $\cG$ as an electrical circuit, one can gain additional insights \cite{klein93,dorfler13,zelazo14,chen16}.
We follow a similar approach when attempting to understand generalized smoothness.
Suppose we interpret an edge $(i,j)$ as a wire between nodes $i$ and $j$, and an edge weight $w_{i,j}$ as \textit{conductance} (equivalently, $1/w_{i,j}$ as the \textit{resistance}) between its two endpoints.
Let $x_i$ and $x_j$ represent the voltage at the two endpoints.
According to \textit{Ohm's law} \cite{robbins12}, the \textit{current} $c_{i,j}$ between the two nodes is the voltage difference at the endpoints times the conductance:
\begin{equation}
c_{i,j} = w_{i,j} (x_i - x_j)
\label{eq:ohm}
\end{equation}

By \textit{Kirchhoff's current law} (KCL) \cite{robbins12}, the net sum of the currents flowing into a node is zero.
Applying KCL to node $2$ in the three-node line graph in Fig.\;\ref{fig:graphEx1} connected by weights $w_{1,2}$ and $w_{2,3}$, we can write:
\begin{equation}
w_{1,2} (x_1 - x_2) + w_{2,3} (x_3 - x_2) = 0
\end{equation}
using Ohm's law (\ref{eq:ohm}).

If we desire a signal $\x$ to satisfy this condition maximally, we can minimize the absolute value of this current sum:
\begin{align}
\min_{\x} & \left|
[\begin{array}{ccc}
-w_{1,2} & (w_{1,2} + w_{2,3}) & -w_{2,3}
\end{array}] ~~ \x
\right| = \left| \L_{2,:}^+ \x \right|
\end{align}
This is in fact the generalized graph-signal smoothness condition we discussed in Section \ref{subsec:gsmooth}.
Thus we can conclude the following: \textit{a graph-signal $\x$ on graph $\cG$ that is perfectly generalized smooth, \textit{i.e.} $|\L^+ \x| = \mathbf{0}$, is a voltage signal on $\cG$ that satisfies KCL}.

This electrical network interpretation also provides an argument why negative edges should not be considered for generalized smoothness.
As done in \cite{dorfler13}, a generalized graph Laplacian $\L_g$ with diagonal element $L_{i,i} \geq \sum_{j|j\neq i} L_{i,j}$ can be considered a conductance matrix, where
an edge $(i,j)$ has branch conductance $-L_{i,j}$ and node $i$ has shunt conductance $L_{i,i} - \sum_{j|j\neq i} L_{i,j} \geq 0$.
For such a resistive circuit, Ohm's law is applicable directly and thus KCL is meaningful.
However, a negative conductance $L_{i,j} < 0$ (equivalently, a negative resistance) means the circuit is no longer resistive, and Ohm's law is not applicable.
As a result, KCL---by extension generalized graph-signal smoothness---is no longer meaningful.

\subsection{Objective Function}
\label{subsec:obj2}

If we choose to include the new generalized smoothness prior to our previous objective (\ref{eq:obj1}) to promote ambiguity in the solution, the objective becomes:

\begin{equation}
\min_{\mathbf{x}} ~
\left\| \mathbf{y} - \mathbf{H} \mathbf{x} \right\|_0
~+~ \mu_1 \; \x^T \, \L_g \, \x
~+~ \mu_2 \; \x^T \, (\L^+)^2 \, \x
\label{eq:obj}
\end{equation}

\subsubsection{Interpretation of Smoothness Priors for Classifiers}

We interpret the two smoothness terms in the context of binary classification.
We know that the true signal $\mathbf{x}$ is indeed \textit{piecewise constant} (PWC); each true label $x_i$ is binary, and labels of the same class cluster together in the same feature subspace.
The graph-signal smoothness term in (\ref{eq:prior}), analogous to the total variation (TV) prior \cite{rudin92} in image restoration \cite{pang17}, promotes a PWC signal $\hat{\x}$ during reconstruction, as empirically demonstrated in previous graph-signal restoration works \cite{liu15,hu16spl,pang17}. 
Hence the smoothness prior is appropriate.

Recall that the purpose of TGV \cite{bredies15} is to avoid over-smoothing a \textit{ramp} (linear increase / decrease in pixel intensity) in an image, which would happen if only a TV prior is used.
A ramp in the reconstructed signal $\hat{\mathbf{x}}$ in our classification context would mean an assignment of label other than $-1$ and $1$, which can reflect the \textit{confidence level} in the estimated label; \textit{e.g.}, a computed label $\hat{x}_i = 0.3$ would mean the classifier has determined that event $i$ is more likely to be $1$ than $-1$, but the confidence level is not high.
We can thus conclude that \textit{the generalized smoothness prior can promote an appropriate amount of ambiguity in the classification solution instead of forcing the classifier to make hard binary decisions.}

\subsection{Iterative Reweighted Least Squares Algorithm}
\label{subsec:IRLS}

To solve (\ref{eq:obj}), we employ the following optimization strategy. We first replace the $l_0$-norm in (\ref{eq:obj}) with a weighted $l_2$-norm:

\begin{small}
\begin{equation}
\min_{\mathbf{x}} ~~
(\y - \H \x)^T \B (\y - \H \x)
~+~ \mu_1 \; \x^T \, \L_g \, \x
~+~ \mu_2 \; \x^T \, (\L^+)^2 \, \x
\label{eq:obj2}
\end{equation}
\end{small}\noindent
where $\B$ is a $K \times K$ diagonal matrix with weights $b_1, \ldots, b_K$ on its diagonal.
In other words, the fidelity term is now a weighted sum of label differences: $(\y - \H \x)^T \B (\y - \H \x) = \sum_{i=1}^{K} b_i (y_i - \H_{i,:} \x)^2$.

The weights $b_i$ should be set so that the weighted $l_2$-norm mimics the $l_0$-norm. To accomplish this, we employ the \textit{iterative reweighted least squares} (IRLS) strategy \cite{daubechies10}, which has been proven to have superlinear local convergence, and solve (\ref{eq:obj2}) iteratively, where the weights $b_i^{(t+1)}$ of iteration $t+1$ is computed using solution $x_i^{(t)}$ of the previous iteration $t$, \textit{i.e.},
\begin{equation}
b_i^{(t+1)} = \frac{1}{(y_i - \H_{i,:}\x^{(t)})^2 + \epsilon}
\label{eq:IRLSweight}
\end{equation}
for a small $\epsilon > 0$ to maintain numerical stability.
Using this weight update, we see that the weighted quadratic term $(\y - \H \x)^T \B (\y - \H \x)$ mimics the original $l_0$-norm $\left\| \y - \H \x \right\|_0$ in the original objective (\ref{eq:obj}) when the solution $\x$ converges.

\subsubsection{Linear System per Iteration}

For a given weight matrix $\B$, it is clear that the objective (\ref{eq:obj2}) is an unconstrained quadratic programming problem with three quadratic terms.
One can thus take the derivative with respect to $\x$ and equating it to zero, resulting in:
\begin{align}
\left( \H^T \B \H + \mu_1 \L_g + \mu_2 (\L^+)^2 \right) \x^* = \H^T \B^T \y
\label{eq:cf}
\end{align}
(\ref{eq:cf}) is a linear system of equations, which can be solved by fast methods like conjugate gradient instead of inverting the sparse matrix on the left.

\subsection{Interpreting Computed Solution $\hat{\mathbf{x}}$}

After the IRLS algorithm converges to a solution $\hat{\mathbf{x}}$, we interpret the classification results as follows.
We perform thresholding by a pre-defined value $\tau$ on $\hat{\mathbf{x}}$ to divide it into three parts, including the rejection option for ambiguous labels:
\begin{equation}
x_i = \begin{cases} 1, & x^*_i > \tau \\
                    \mathrm{Rejection}, & -\tau<x^*_i<\tau \\
                    -1, & x^*_i < -\tau.
      \end{cases}
\label{eq:classify}
\end{equation}
Typically, the fraction of tolerable rejection labels is set per application requirement. 
Clearly, eliminating more ambiguous labels leads to a larger tolerable rejection rate, resulting in a smaller classification error rate.

\section{Experimentation}
\label{sec:results}
\subsection{Experiment Setup}

\subsubsection{Datasets for Training and Testing}

To evaluate the performances of different classification methods, we selected four two-class datasets from the KEEL (Knowledge Extraction based on Evolutionary Learning) database \cite{keel09}, which contains a rich collection of labeled and unlabeled datasets for data mining and analysis and face gender dataset provided in \cite{uk09}.

The first dataset is the Phoneme dataset that provides values of five categorical attributes to distinguish nasal sounds (class 0) from oral sounds (class 1).
The second is the Banana dataset, an artificial dataset where 5300 instances belong to several clusters with a banana shape. In the dataset, two attributes were extracted to classify two kinds of banana shapes.
The third is the Face Gender dataset that consists of 7900 face images (395 individuals, 20 images per individual). We extract the Local Binary Pattern (LBP) features to represent the faces for classifying the genders of faces.
The fourth dataset called ``Sonar, Mines vs. Rocks" contains various patterns obtained by bouncing sonar signals off metal cylinders and rocks at various angles and under various conditions. Each pattern is a set of 60 numbers in the range $[0.0, 1.0]$, where each number represents the energy within a particular frequency band, integrated over a certain period of time. These patterns are used to classify an object to a metal cylinder or a rock.

For our experiments, we randomly sampled 300 instances from the first and second dataset, 400 instances from the third and 210 instances from the fourth, and used 70\% of the samples as training data and 30\% as testing data. We repeated the process 100 times for each dataset and then calculated the average performance of the 100 experiments in terms of classification error rate.

\subsubsection{Graph Construction}

To construct a graph for our proposed methods, we first constructed an initial graph with positive edge weights.
For each sample (node), we found its three nearest neighbors according to the Euclidean distances between the node and its neighbors, and connected these nodes using edges with positive weights that are normalized to [0,1] using the Gaussian kernel in (\ref{eq:edgeWeight}).
We performed clustering on the labeled nodes in the graph and found the centroids and boundaries of the two clusters, as explained in Section\;\ref{subsec:construct}.
We then assigned a negative edge weight between each pair with a value normalized to [$-10$,0], [$-20$,0], [$-1$,0] and [$-1$,0] for the four datasets respectively, where the magnitude is proportional to the Euclidean distance between the pair.
For boundaries, We paired the cluster boundaries based on feature distances to find the boundary samples of the two clusters and assigned a negative edge weight between each pair with a value normalized to [$-0.01$,0], [$-0.01$,0], [$-0.01$,0] and [$-0.1$,0] for the four datasets respectively, where the magnitude is proportional to the Euclidean distance between the pair.

\subsubsection{Comparison Schemes}

We tested our proposed algorithm against eight schemes: i) linear SVM, ii) SVM with a RBF kernel (named SVM-RBF), iii) a more robust version of the famous AdaBoost called RobustBoost \cite{RobustBoost09} that claims robustness against label noise, iv) a graph classifier with the graph-signal smoothness prior (\ref{eq:smoothness}) where the edge weights of the graph are all positive (named Graph-Pos), v) a graph classifier with a graph containing negative edge weights where the graph Laplacian $\L$ is perturbed by the minimum-norm perturbation criteria in (\ref{eq:minNorm}) to eliminate negative eigenvalues for numerical stability (named Graph-MinNorm), vi) a bandlimited graph method proposed in \cite{sun13} (named Graph-Bandlimited), vii) a graph classifier using a smoothness prior based on the adjacency matrix proposed in \cite{chen15} (named Graph-AdjSmooth), and viii) a semi-supervised learning  algorithm based on graph wavelet \cite{ekambaram13} (named Graph-Wavelet).

We implemented four variants of our proposed minimum-variance perturbation graph classifier. The first three utilize the generalized graph Laplacian $\L_g$ without the generalized smoothness term (\textit{i.e.}, $\mu_2=0$ in (\ref{eq:obj}) and $\tau=0$ in (\ref{eq:classify})) based on three different negative edge weights assignment schemes as described in Section \ref{subsec:construct}: i) assigning negative edge weights between the centroid sample pairs (named Proposed-Centroid); ii) assigning negative edge weights between boundary sample pairs (named Proposed-Boundary); and iii) assigning negative edge weights between the centroid sample pairs and between the boundary sample pairs (named Proposed-Hybrid). The fourth variant is the proposed method in (\ref{eq:obj}) with rejection (named Proposed-Rej) where the rejection rate is controlled to be within 9--10\% by tuning parameters in (\ref{eq:classify}).

\subsection{Performance Evaluation}

\begin{table}
\caption{Classification error rates in the Phoneme dataset for competing schemes under different training label error rates (the numbers in the parentheses of the last row indicate the rejection rates)}
\label{tab:Phoneme}
\vspace{-0.15in}
\begin{center}
\begin{scriptsize}
\begin{tabular}{|c|c|c|c|c|c|} \hline
  \% label noise & 0\% & 5\% & 10\% & 15\%  & 20\% \\ \hline
SVM-Linear    &    21.83\% & 23.35\% & 24.55\% & 25.05\% & 25.64\% \\ \hline
SVM-RBF       &    16.63\% & 16.84\% & 17.48\% & 17.72\% & 19.34\% \\ \hline
RobustBoost \cite{RobustBoost09} &    12.81\% & 14.91\% & 17.94\% & 19.33\% & 21,50\% \\ \hline
Graph-Pos     &    13.22\% & 14.91\% & 16.79\% & 18.17\% & 20.70\% \\ \hline
Graph-MinNorm &    12.90\% & 14.53\% & 16.58\% & 18.45\% & 20.56\% \\ \hline
Graph-Bandlimited \cite{sun13} &    11.70\% & 14.06\% & 17.05\% & 18.70\% & 21.29\% \\ \hline
Graph-AdjSmooth \cite{chen15} &    11.31\% & 13.69\% & 16.79\% & 18.65\% & 20.67\% \\ \hline
Graph-Wavelet \cite{ekambaram13} &    27.25\% & 28.84\% & 30.48\% & 31.95\% & 33.51\% \\ \hline
Proposed-Centroid &    10.81\% & 13.09\% & 16.18\% & 17.87\% & 20.47\% \\ \hline
Proposed-Boundary &    12.14\% & 14.44\% & 17.18\% & 19.02\% & 21.51\% \\ \hline
Proposed-Hybrid      &    \textbf{10.57\%} & \textbf{13.00\%} & \textbf{15.44\%}  & \textbf{17.14\%} & \textbf{19.15\% }\\ \hline
\multirow{2}{*}{Proposed-Rej}  & 9.85\% & 11.53\% & 13.97\%  & 14.96\% & 17.03\%  \\ & (9.44\%) & (9.69\%) &  (9.46\%)  &  (9.81\%)&   (9.80\%) \\ \hline
\end{tabular}
\end{scriptsize}
\vspace{-0.15in}
\end{center}
\end{table}

\begin{table}
\caption{Classification error rates in the Banana dataset for competing schemes under different training label error rates (the numbers in the parentheses of the last row indicate the rejection rates)}
\label{tab:Banana}
\vspace{-0.15in}
\begin{center}
\begin{scriptsize}
\begin{tabular}{|c|c|c|c|c|c|} \hline
  \% label noise & 0\% & 5\% & 10\% & 15\%  & 20\% \\ \hline
SVM-Linear        &    54.71\% & 54.97\% & 54.70\%  & 53.95\% & 53.42\% \\ \hline
SVM-RBF           &    12.49\% & 13.27\% & 13.72\%  & 16.23\% & 18.63\% \\ \hline
RobustBoost \cite{RobustBoost09} &    20.42\% & 22.73\% & 24.53\%  & 25.12\% & 27.52\% \\ \hline
Graph-Pos         &    14.05\% & 15.89\% & 18.02\%  & 20.76\% & 21.93\% \\ \hline
Graph-MinNorm     &    10.23\% & 12.37\% & 14.44\%  & 17.41\% & 18.69\% \\ \hline
Graph-Bandlimited \cite{sun13} &    7.53\% & 11.77\% & 15.80\%  & 19.14\% & 21.07\% \\ \hline
Graph-AdjSmooth \cite{chen15} &    8.85\% & 12.08\% & 15.28\%  & 18.26\% & 20.67\% \\ \hline
Graph-Wavelet \cite{ekambaram13}  &    23.18\% & 24.25\% & 25.70\%  & 27.15\% & 30.13\% \\ \hline
Proposed-Centroid &    \textbf{5.17\%} & 10.50\% & 13.79\%  & 16.80\% & 19.39\% \\ \hline
Proposed-Boundary &    13.37\% & 15.68\% & 18.27\%  & 20.51\% & 22.72\% \\ \hline
Proposed-Hybrid          &    5.36\% & \textbf{9.43\%} & \textbf{12.79\%}  & \textbf{16.04\%} & \textbf{18.43\%} \\ \hline
\multirow{2}{*}{Proposed-Rej}   &   3.74\% & 6.57\% & 9.26\% & 12.19\% & 14.06\% \\ & (9.59\%)  & (9.89\%) & (9.14\%) &  (9.96\%) & (9.95\%) \\ \hline
\end{tabular}
\end{scriptsize}
\vspace{-0.15in}
\end{center}
\end{table}

\begin{table}
\caption{Classification error rates in the Face Gender dataset for competing schemes under different training label error rates (the numbers in the parentheses of the last row indicate the rejection rates)}
\label{tab:Gender}
\vspace{-0.15in}
\begin{center}
\begin{scriptsize}
\begin{tabular}{|c|c|c|c|c|c|} \hline
  \% label noise & 0\% & 5\% & 10\% & 15\%  & 20\% \\ \hline
SVM-Linear        &    17.65\% & 18.22\% &18.77\% & 19.59\% & 21.6\% \\ \hline
SVM-RBF           &    12.14\% & 12.16\% & 12.83\% & 16.30\% & 24.01\% \\ \hline
RobustBoost \cite{RobustBoost09} &    9.15\% & 11.09\% & 14.36\% & 17.36\% & 20.68\% \\ \hline
Graph-Pos         &    13.15\% & 13.62\% & 14.38\% & 15.39\% & 16.54\% \\ \hline
Graph-MinNorm     &    7.15\% & 8.26\% & 9.48\% & 10.37\% & 12.01\% \\ \hline
Graph-Bandlimited \cite{sun13} &    5.78\% & 11.83\% & 15.30\% & 19.74\% & 23.44\% \\ \hline
Graph-AdjSmooth \cite{chen15}  &    \textbf{1.25\%} & 5.01\% & 7.94\% & 11.45\% & 15.39\% \\ \hline
Graph-Wavelet \cite{ekambaram13}  &    20.02\% & 19.95\% & 20.12\% & 20.7\% & 21.43\% \\ \hline
Proposed-Centroid &    1.44\% & \textbf{2.96\%} & 4.46\% & 5.88\% & 8.07\% \\ \hline
Proposed-Boundary &    10.81\% & 12.09\% & 13.17\% & 14.33\% & 15.96\% \\ \hline
Proposed-Hybrid      &    1.71\% & 3.02\% & \textbf{4.22\%}  & \textbf{5,75\%} & \textbf{7.71\%} \\ \hline
\multirow{2}{*}{Proposed-Rej} &    0.36\% & 0.68\% & 1.08\% & 2.39\% & 4.18\% \\ & (9.70\%) & (9.29\%) & (9.85\%) & (9.08\%) & (9.05\%) \\ \hline
\end{tabular}
\end{scriptsize}
\vspace{-0.15in}
\end{center}
\end{table}

\begin{table}
\caption{Average classification error rates in the Sonar, Mines vs. Rocks dataset for competing schemes under different training label error rates and three different $\sigma_0$ weights (the numbers in the parentheses of the last row indicate the rejection rates)}
\label{tab:Sonar}
\vspace{-0.15in}
\begin{center}
\begin{scriptsize}
\begin{tabular}{|c|c|c|c|c|c|} \hline
  \% label noise & 0\% & 5\% & 10\% & 15\%  & 20\% \\ \hline
Graph-Pos        &    22.62\% & 24.28\% & 26.07\% & 28.63\% & 30.27\% \\ \hline
Graph-MinNorm    &    21.89\% & 23.53\% & 25.61\% & 28.14\% & 30.04\% \\ \hline
Graph-AdjSmooth \cite{chen15}  &    24.20\% & 25.56\% & 26.76\% & 27.91\% & 30.82\% \\ \hline
Proposed-Hybrid  &    \textbf{19.42\%} & \textbf{20.99\%} & \textbf{22.36\%}  & \textbf{24.30\%} & \textbf{25.93\% }\\ \hline
\end{tabular}
\end{scriptsize}
\vspace{-0.15in}
\end{center}
\end{table}

To evaluate the robustness of different classification schemes against label noise, we randomly selected a portion of samples from the training set and reversed their labels.
All the classifiers were then trained using the same set of features and labels.
Each test set was classified by the classifiers and the results are compared with the ground-truth labels.

\begin{table}
\caption{Classification error and rejection rates in the Banana dataset for the proposed method (with rejection) under different training label error rates and $\sigma_1$ (the numbers in the parentheses of the last row indicate the rejection rates)}
\label{tab:ErrorRejection}
\vspace{-0.15in}
\begin{center}
\begin{scriptsize}
\begin{tabular}{|c|c|c|c|c|c|} \hline
  \% label noise & 0\% & 5\% & 10\% & 15\%  & 20\% \\ \hline
Proposed-Rej   &    3.74\%  & 6.57\%  & 9.26\%   & 12.19\%  & 14.06\%  \\
($\sigma_1=1$)  &   (9.59\%)  & (9.89\%) & (9.14\%)  & (9.96\%) & (9.95\%) \\ \hline
Proposed-Rej   &    3.57\%  & 6.56\%  & 9.21\%   & 12.04\%  & 14.06\%  \\
($\sigma_1=0.8$)  & (10.63\%)  & (9.92\%) & (9.21\%)  & (10.06\%) & (9.95\%) \\ \hline
Proposed-Rej   &    3.46\%   & 6.50\%  & 9.14\%  & 12.05\%  & 14.00\%  \\
($\sigma_1=0.6$)  &   (12.56\%)  &  (10.21\%) & (9.42\%)  & (10.06\%) &  (10.09\%) \\ \hline
\end{tabular}
\end{scriptsize}
\vspace{-0.15in}
\end{center}
\end{table}

\subsubsection{Numerical comparisons for different label noise}

The resulting classification error rates for the first three datasets using different classifiers are presented in Tables \ref{tab:Phoneme}--\ref{tab:Gender}, where the percentage of randomly erred training labels ranges from 0\% to 20\%.
The comparisons show that our proposed scheme achieves the lowest classification error rates when compared to the competing schemes under almost all training label error rates.
The parameters $(\gamma, \sigma_0, \sigma_1, \tau)$ used for the three datasets respectively are:
$(1, 1, 1, [0.0115, 0.027])$,
$(1, 0.1, 1, [0.000055, 0.00035])$,
$(1, 0.1, 2, [0.0095, 0.025])$.
Compared to the graph classifiers Graph-Pos and Graph-AdjSmooth, our results show that adding negative edge weights can effectively improve the classification accuracy by 1.03--8.5\% and 0.54--7.68\%, respectively.
We can also observe that our proposed matrix perturbation scheme significantly outperforms the minimum-norm based perturbation (Graph-MinNorm) in classification accuracy. Compared to Graph-Bandlimited and Graph-Wavelet, the proposed hybrid method improves the classification accuracy by 1.06--15.73\% and 11.11--19.21\%, respectively.

Further, as shown in Table\;\ref{tab:Sonar}, we evaluate the performances of those four graph classifiers that employ a smoothness prior (i.e., Graph-Pos, Graph-AdjSmooth, Graph-MinNorm, and Proposed-Hybrid) using a range of weight parameter values $\mu_1$. For the first three methods, we set  $\mu_1$ to be 10, 1, and 0.1, respectively, and then calculate the average classification error rate accordingly, whereas for Proposed-Hybrid, we set the value of $\mu_1$ to be 0.1, 0.01, and 0.001, respectively. Table\;\ref{tab:Sonar} shows that the proposed method is not sensitive to the change of $\mu_1$ and, compared to the other three graph classifiers, improves the classification accuracy by 2.47--4.89\% for the ``Sonar, Mines vs. Rocks" dataset.

\subsubsection{Graph classifier with non-zero rejection rate}

By allowing a certain amount of ambiguous samples to remain unlabeled (less than 10\% rejection rate in our experiments), our proposed generalized graph-signal smoothness prior can further improve the classification accuracy.
We note that a user may define the desired classifier performance as a weighted sum of classification error and rejection rate for different applications, as done in  \cite{chow70}.
Table \ref{tab:ErrorRejection} shows the classification error and rejection rates in the ``Banana" dataset for our proposed method with rejection under different training label error rates and $\mu_2$, where the values of $\tau$ are set the same as that used in the first row.
It shows that as $\mu_2$ for the generalized smoothness term increases, the rejection rate also increases, which is consistent with our explanation in Section \ref{subsec:gsmooth} that the second smoothness term promotes ambiguity in the solution instead of forcing the solution to be strictly binary.
As a result, using our algorithm, one can thus tune $\mu_2$ and $\tau$ to adjust the preference of classification error versus rejection rate.

\begin{figure}

\begin{minipage}[b]{.95\linewidth}
 \centering
 \centerline{\includegraphics[width=9.2cm]{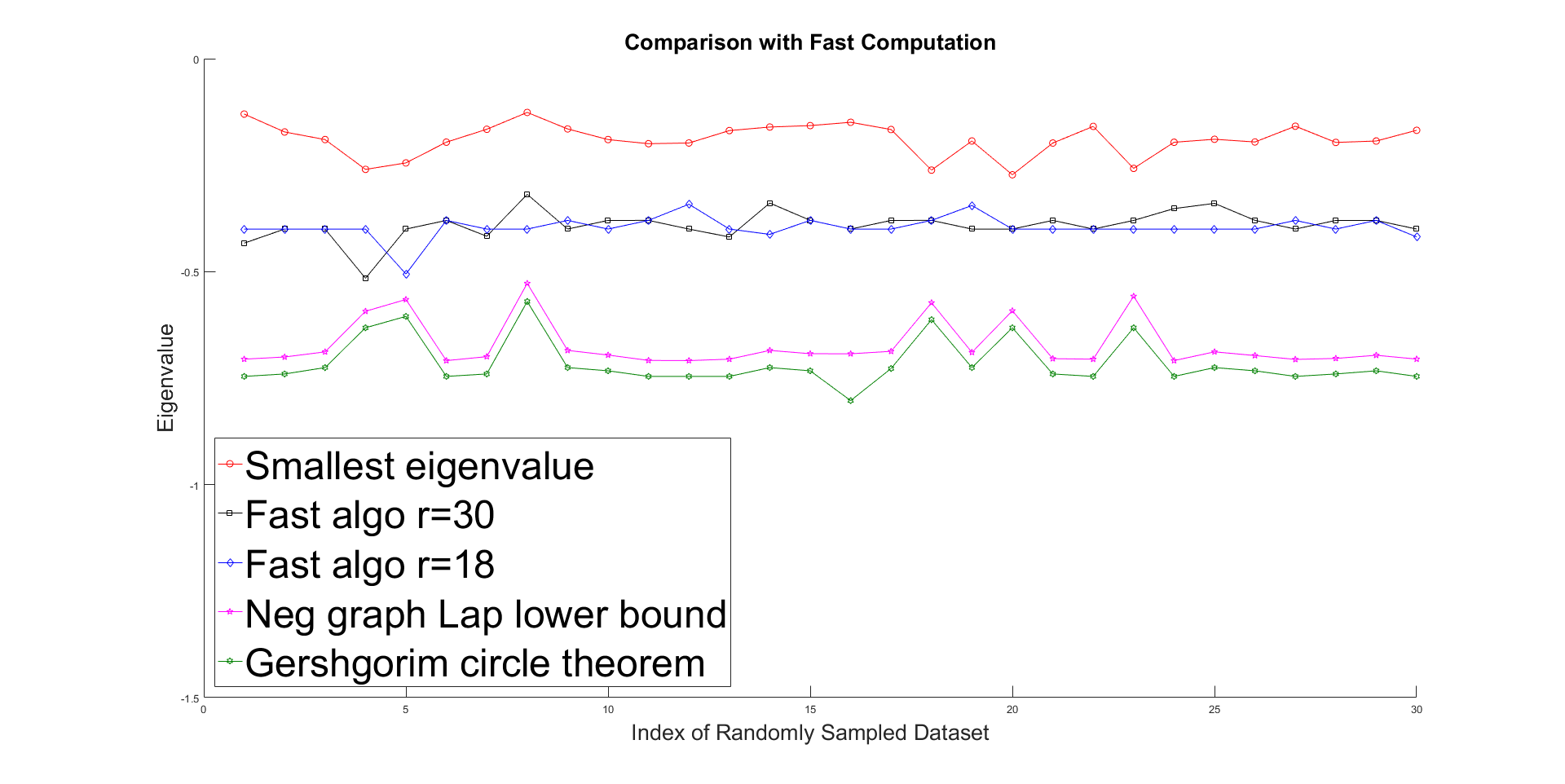}}
\end{minipage}

\vspace{-0.15in}
\caption{\small{Comparison of the actual smallest eigenvalues and their lower-bounds using  $r$ = 30 and 18 for the Phoneme dataset ($N = 300$), corresponding to 99\% and 99.64\% computation reduction.}}
\label{fig:FastMines}
\end{figure}

\begin{figure}

\begin{minipage}[b]{.95\linewidth}
 \centering
 \centerline{\includegraphics[width=9.2cm]{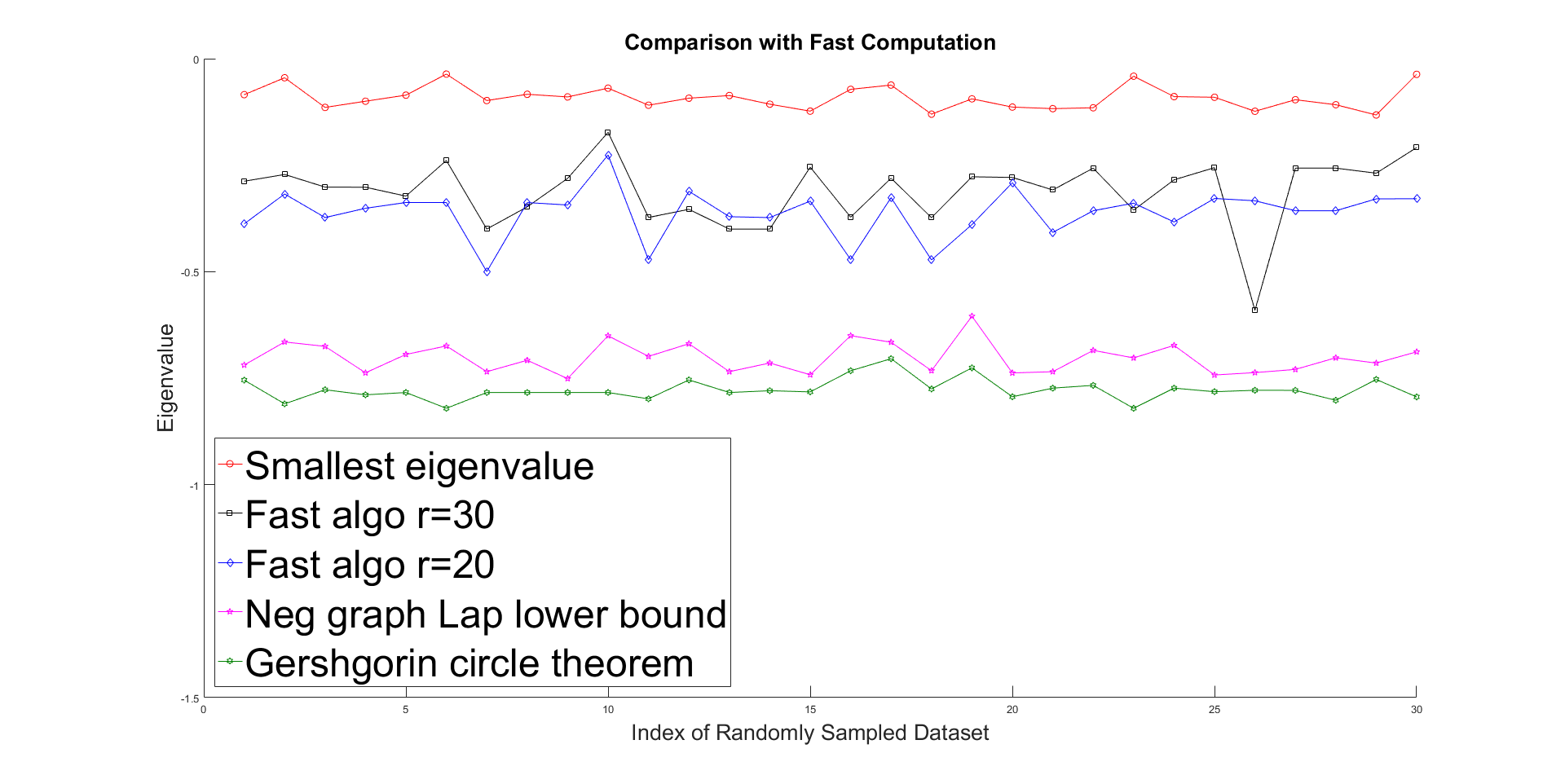}}
\end{minipage}

\vspace{-0.15in}
\caption{\small{Comparison of the actual smallest eigenvalues and their lower-bounds using $r$ = 30 and 20 for the gender dataset ($N = 400$), corresponding to 99\% and 99.75\%, computation reduction, respectively.}}
\label{fig:FastBanana}
\end{figure}

\begin{table}
\caption{Classification error rates in the three datasets for the proposed minimum-variance matrix perturbation method with actual minimum and lower-bound eigenvalues under different training label error rates}
\label{tab:Fast}
\vspace{-0.15in}
\begin{center}
\begin{scriptsize}
\begin{tabular}{|c|c|c|c|c|c|} \hline
  \% label noise & 0\% & 5\% & 10\% & 15\%  & 20\% \\ \hline
\multicolumn{6}{ |c| }{Phoneme dataset} \\ \hline
Proposed        &    10.57\% & 13.00\% & 15.44\%  & 17.14\% & 19.15\% \\ \hline
Fast ($r=150$)  &    10.61\% & 13.04\% & 15.35\%  & 17.22\% & 19.23\% \\ \hline
Fast ($r=100$)  &    10.60\% & 13.06\% & 15.47\%  & 17.19\% & 19.26\% \\ \hline
\multicolumn{6}{ |c| }{Banana dataset} \\ \hline
Proposed        &    5.36\% & 9.43\% & 12.79\%  & 16.04\% & 18.43\% \\ \hline
Fast ($r=150$)  &    5.44\% & 9.44\% & 12.59\%  & 15.87\% & 18.15\% \\ \hline
Fast ($r=100$)  &    5.31\% & 9.42\% & 12.73\%  & 15.81\% & 18.31\% \\ \hline
\end{tabular}
\end{scriptsize}
\vspace{-0.15in}
\end{center}
\end{table}

\subsubsection{Fast computation}

In Section\;\ref{subsec:LowerBound}, we proposed a fast eigen-decomposition scheme to lower-bound the smallest eigenvalue $\lambda_{\min}$ of the graph Laplacian $\L$ for a graph with negative edge weights.
Fig.\;\ref{fig:FastMines} and Fig.\;\ref{fig:FastBanana} show the actual minimum eigenvalues (denoted Smallest eigenvalue), their computed lower-bounds (denoted Fast), negative graph Laplacian $\L^-$ lower bound $-\lambda^-_{\max}$ (\ref{eq:lbound}) and lower-bounds computed using the Gershgorin circle theorem \cite{gewiki10} in the first 30 out of 100 sampled data subsets for two datasets, respectively.
In the experiment, we set parameter $r$ to be about $\sqrt{N}$ and 30 of total number of samples $N$, which correspond to about 99\% and 99.64-99.75\% computation reduction, respectively, since the computation complexity is reduced from $O(N^3)$ to $O(N r^2)$ as explained in Section\;\ref{subsec:complexity}.

The results show that $\lambda_{\min}^{\#}$ computed by our proposed fast algorithm is an actual lower-bound for the true minimum eigenvalue $\lambda_{\min}$, \textit{i.e.} $\lambda_{\min}^{\#} \leq \lambda_{\min}$. 
$\lambda_{\min}^{\#}$ is also a tighter lower bound than the two alternatives computed using negative graph Laplacian and the Gershgorin circle theorem.
The proposed fast algorithm then obtains $\L_g$ using matrix perturbation $-\lambda_{\min}^{\#} \mathbf{I}$.
Table\;\ref{tab:Fast} compares the classification error rates of the proposed perturbation method (without rejection) using the actual minimum eigenvalues with the approximation computing the lower-bound eigenvalues by our proposed fast algorithm.
Results show that the fast algorithm leads to slight performance differences compared to the full computation method.

\begin{figure}

\begin{minipage}[b]{.95\linewidth}
 \centering
 \centerline{\includegraphics[width=9.2cm]{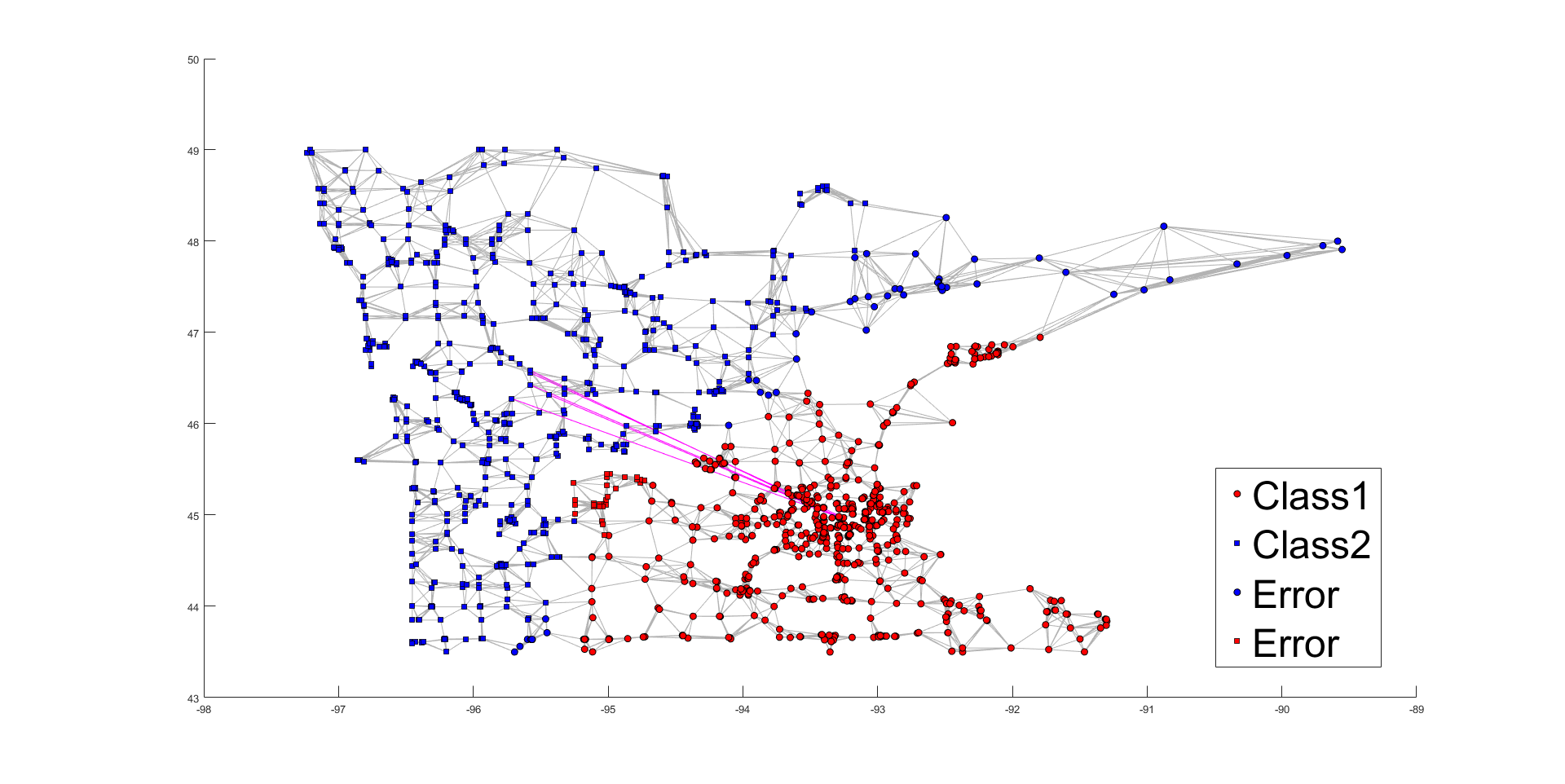}}
\end{minipage}

\vspace{-0.15in}
\caption{\small{Visualization of classification result in graph based on first eigenvector result for Minnesota road network dataset, the purple lines are negative edges.}}
\label{fig:firsteig}
\end{figure}

\begin{figure}

\begin{minipage}[b]{.95\linewidth}
 \centering
 \centerline{\includegraphics[width=9.2cm]{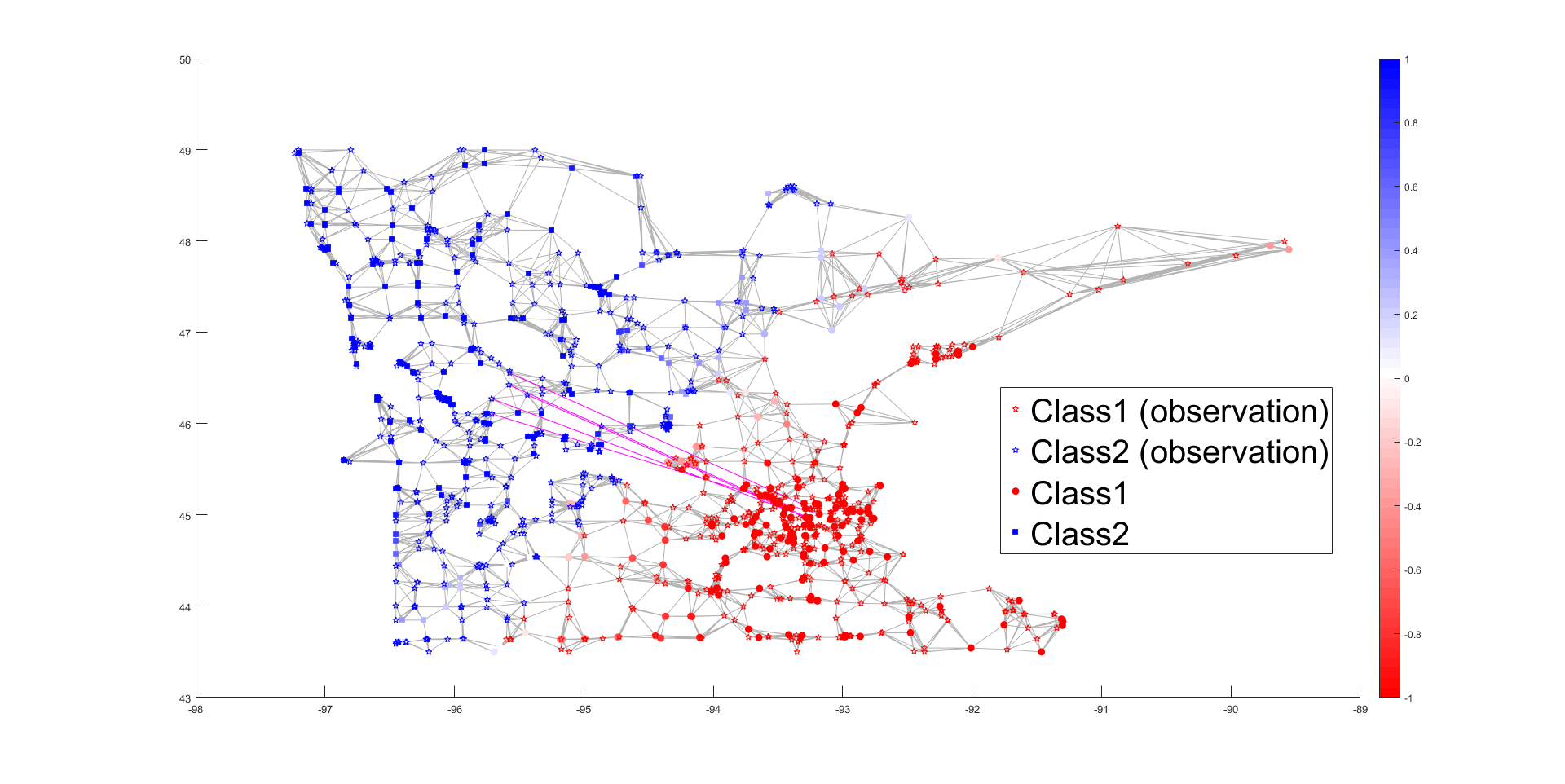}}
\end{minipage}

\vspace{-0.15in}
\caption{\small{Visualization of reconstruction of x in graph for Minnesota road network dataset (the deeper the color is, the reconstruction of x is closer to 1 or -1), the purple lines are negative edges.}}
\label{fig:reconstructionX}
\end{figure}
\subsubsection{Visualization result}

In Section\;\ref{subsec:graphexample}, we use a simple example to illustrate why by using negative edge weights, the resulting low graph frequency components of an indefinite graph Laplacian $\L$ can be useful in restoring signal $\x$.
In this subsection, we use the Minnesota road network dataset that provides 2642 $x$- and $y$- coordinates with road network data and randomly sampled 1400 instances to construct a larger and more complex graph.
We apply centroid-based method and assign a negative edge weight between each pair with a value normalized to [$-1$,0] based on the Euclidean distance between the pair.
Fig.\;\ref{fig:firsteig} shows the first eigenvector of the graph Laplacian with negative edges, which  reflects different class labels.
Further, we show the reconstructed $\x$ in Fig.\;\ref{fig:reconstructionX} where the deeper the color, the closer the reconstructed sample is to 1 or -1.


\section{Conclusion}
\label{sec:conclude}
To address the semi-supervised learning problem, in this paper we view a classifier as a graph-signal in a high-dimensional feature space, and pose a maximum a posteriori (MAP) problem to restore the classifier signal given partial and noisy labels.
Unlike previous graph-based classifier works, we consider in addition edges with negative weights that signify dissimilarity between sample pairs.
To achieve a stable signal smoothness prior, we derive a minimum-norm perturbation matrix $\Del$ that preserves the original eigen-structure, so that when added to the graph Laplacian $\L$, the matrix sum is positive semi-definite (PSD). 
We can compute a fast approximation to $\Del$ using a recursive algorithm based on the Haynsworth inertia additivity formula.
Finally, we show that a generalized smoothness prior can promote ambiguity in the classifier signal, so that estimated labels with low confidence can be rejected.
Experimental results show that our proposal outperforms SVM variants and previous graph-based classifiers using positive-edge graphs noticeably. 

\section*{Acknowledgment}
The authors thank the authors of \cite{ekambaram13} for providing source code of the graph-wavelet-based  semi-supervised learning algorithm. 
The authors thank Prof. Michael Ng of Hong Kong Baptist University for the valuable discussion on an early draft of this paper. 



\appendix

We prove that if $\L_{1,1}$ is PD and $\L/\L_{1,1} + \delta \L$ is PSD for $\delta > 0$, then given perturbed matrix $\L' = \L + \delta \I$, $\L'/\L_{1,1}'$ is also PSD.
By definition, $\L / \L_{1,1} + \delta \I$ is PSD means:
\begin{align}
\x^T \left( \L_{2,2} - \L_{1,2}^T \L_{1,1}^{-1} \L_{1,2} + \delta \I \right) \x \geq 0 \nonumber 
\end{align}
Let $\L_{1,1}$ be spectrally decomposed to $\L_{1,1} = \V \Lamb \V^T$, where $\Lamb = \textrm{diag}(\lambda_1, \ldots, \lambda_n)$ is a diagonal matrix containing all positive eigenvalues, since $\L_{1,1}$ is PD.
We can thus rewrite above:
\begin{align}
& \x^T \L_{2,2} \x - 
\underbrace{\x^T \L_{1,2}^T \V}_{\y^T} \textrm{diag}(\lambda_1^{-1}, \ldots, \lambda_n^{-1}) \underbrace{\V^T \L_{1,2} \x}_{\y} + \delta \x^T \x
\nonumber \\
& = \x^T \L_{2,2} \x + \delta \x^T \x - \sum_{i} \lambda_i^{-1} y_i^2
\label{eq:append1}
\end{align}  

If $\L$ is now perturbed by $\delta \I$, we can similarly write the resulting SC $\L'/\L_{1,1}'$ in quadratic form:
\begin{align}
\x^T \left( \L_{2,2} + \delta \I - \L_{1,2}^T (\L_{1,1} + \delta \I)^{-1} \L_{1,2} \right) \x 
\nonumber \\
= \x^T \L_{2,2} \x + \delta \x^T \x - \x^T \L_{1,2}^T (\L_{1,1} + \delta \I)^{-1} \L_{1,2} \x
\label{eq:append2}
\end{align}
The first two terms are the same as ones in (\ref{eq:append1}).
The third term can be rewritten as:
\begin{align}
& \underbrace{\x^T \L_{1,2}^T \V}_{\y^T} 
\mathrm{diag}((\lambda_1 + \delta)^{-1}, \ldots, (\lambda_n + \delta)^{-1}) 
\underbrace{\V^T \L_{1,2} \x}_{\y} \nonumber \\
& = \sum_i (\lambda_i + \delta)^{-1} y_i^2
\end{align}
Since $\lambda_i > 0$, we see that $\lambda_i^{-1} > (\lambda_i + \delta)^{-1}$.
Hence this third term has magnitude strictly smaller than one in (\ref{eq:append1}).
Thus, non-negativity in (\ref{eq:append1}) implies non-negativity in (\ref{eq:append2}). 
Thus $\L'/\L_{1,1}'$ is PSD. $\Box$

\bibliographystyle{IEEEtran}
\bibliography{./IEEEabrv_mod,ref2}

\begin{thebibliography}{10}
\providecommand{\url}[1]{#1}
\csname url@samestyle\endcsname
\providecommand{\newblock}{\relax}
\providecommand{\bibinfo}[2]{#2}
\providecommand{\BIBentrySTDinterwordspacing}{\spaceskip=0pt\relax}
\providecommand{\BIBentryALTinterwordstretchfactor}{4}
\providecommand{\BIBentryALTinterwordspacing}{\spaceskip=\fontdimen2\font plus
\BIBentryALTinterwordstretchfactor\fontdimen3\font minus
  \fontdimen4\font\relax}
\providecommand{\BIBforeignlanguage}[2]{{%
\expandafter\ifx\csname l@#1\endcsname\relax
\typeout{** WARNING: IEEEtran.bst: No hyphenation pattern has been}%
\typeout{** loaded for the language `#1'. Using the pattern for}%
\typeout{** the default language instead.}%
\else
\language=\csname l@#1\endcsname
\fi
#2}}
\providecommand{\BIBdecl}{\relax}
\BIBdecl

\bibitem{bishop06}
C.~Bishop, \emph{Pattern Recognition and Machine Learning}.\hskip 1em plus
  0.5em minus 0.4em\relax Springer, 2006.

\bibitem{zhou03}
D.~Zhou, O.~Bousquet, T.~N. Lal, J.~Weston, and B.~Scholkopf, ``Learning with
  local and global consistency,'' in \emph{16th International Conference on
  Neural Information Processing (NIPS)}, Whistler, Canada, December 2003.

\bibitem{belkin04}
M.~Belkin, I.~Matveeva, and P.~Niyogi, ``Regularization and semi-supervised
  learning on large graphs,'' in \emph{Shawe-Taylor J., Singer Y. (eds)
  Learning Theory, COLT 2004, Lecture Notes in Computer Science}, vol.
  3120.\hskip 1em plus 0.5em minus 0.4em\relax Springer, Berlin, Heidelberg,
  2004, pp. 624--638.

\bibitem{gavish10}
M.~Gavish, B.~Nadler, and R.~Coifman, ``Multiscale wavelets on trees, graphs
  and high dimensional data: Theory and applications to semi supervised
  learning,'' in \emph{27th International Conference on Machine Learning},
  Haifa, Israel, June 2010.

\bibitem{shuman11}
D.~Shuman, M.~Faraji, and P.~Vandergheynst, ``Semi-supervised learning with
  spectral graph wavelets,'' in \emph{International Conference on Sampling
  Theory and Applications (SampTA)}, Singapore, May 2011.

\bibitem{ekambaram13}
V.~Ekambaram, G.~Fanti, B.~Ayazifar, and K.~Ramchandran, ``Wavelet-regularized
  graph semi-supervised learning,'' in \emph{Symposium on Graph Signal
  Processing in IEEE Global Conference on Signal and Information Processing
  (GlobalSIP)}, Austin, TX, December 2013.

\bibitem{guillory09}
A.~Guillory and J.~Bilmes, ``Label selection on graphs,'' in \emph{Twenty-Third
  Annual Conference on Neural Information Processing Systems}, Vancouver,
  Canada, December 2009.

\bibitem{zhang14}
L.~Zhang, C.~Cheng, J.~Bu, D.~Cai, X.~He, and T.~Huang, ``Active learning based
  on locally linear reconstruction,'' in \emph{IEEE Transactions on Pattern
  Analysis and Machine Intelligence}, vol. 33, no.10, October 2014, pp.
  2026--2038.

\bibitem{chen15}
S.~Chen, A.~Sandryhaila, J.~Moura, and J.~Kovacevic, ``Signal recovery on
  graphs: Variation minimization,'' in \emph{IEEE Transactions on Signal
  Processing}, vol. 63, no.17, September 2015, pp. 4609--4624.

\bibitem{gadde14}
A.~Gadde, A.~Anis, and A.~Ortega, ``Active semi-supervised learning using
  sampling theory for graph signals,'' in \emph{ACM SIGKDD International
  Conference on Knowledge Discovery and Data Mining}, New York, NY, August
  2014.

\bibitem{chung96}
F.~Chung, \emph{Spectral Graph Theory}.\hskip 1em plus 0.5em minus 0.4em\relax
  American Mathematical Society, 1996.

\bibitem{shuman13}
D.~I. Shuman, S.~K. Narang, P.~Frossard, A.~Ortega, and P.~Vandergheynst, ``The
  emerging field of signal processing on graphs: Extending high-dimensional
  data analysis to networks and other irregular domains,'' in \emph{IEEE Signal
  Processing Magazine}, vol. 30, no.3, May 2013, pp. 83--98.

\bibitem{mao16}
Y.~Mao, G.~Cheung, C.-W. Lin, and Y.~Ji, ``Image classifier learning from noisy
  labels via generalized graph smoothness priors,'' in \emph{IEEE IVMSP
  Workshop}, Bordeaux, France, July 2016.

\bibitem{hu13}
W.~Hu, X.~Li, G.~Cheung, and O.~Au, ``Depth map denoising using graph-based
  transform and group sparsity,'' in \emph{IEEE International Workshop on
  Multimedia Signal Processing}, Pula, Italy, October 2013.

\bibitem{narang13}
S.~K. Narang, A.~Gadde, E.~Sanou, and A.~Ortega, ``Localized iterative methods
  for interpolation in graph structured data,'' in \emph{Symposium on Graph
  Signal Processing in IEEE Global Conference on Signal and Information
  Processing (GlobalSIP)}, Austin, TX, December 2013.

\bibitem{narang13icassp}
S.~K. Narang, A.~Gadde, and A.~Ortega, ``Signal processing techniques for
  interpolation of graph structured data,'' in \emph{IEEE International
  Conference on Acoustics, Speech and Signal Processing}, Vancouver, Canada,
  May 2013.

\bibitem{mao16tmm}
Y.~Mao, G.~Cheung, and Y.~Ji, ``On constructing $z$-dimensional
  {DIBR}-synthesized images,'' in \emph{IEEE Transactions on Multimedia}, vol.
  18, no.8, August 2016, pp. 1453--1468.

\bibitem{wan16}
P.~Wan, G.~Cheung, D.~Florencio, C.~Zhang, and O.~Au, ``Image bit-depth
  enhancement via maximum-a-posteriori estimation of {AC} signal,'' in
  \emph{IEEE Transactions on Image Processing}, vol. 25, no.6, June 2016, pp.
  2896--2909.

\bibitem{liu15}
X.~Liu, G.~Cheung, X.~Wu, and D.~Zhao, ``Inter-block soft decoding of {JPEG}
  images with sparsity and graph-signal smoothness priors,'' in \emph{IEEE
  International Conference on Image Processing}, Quebec City, Canada, September
  2015.

\bibitem{hu16spl}
W.~Hu, G.~Cheung, and M.~Kazui, ``Graph-based dequantization of
  block-compressed piecewise smooth images,'' in \emph{IEEE Signal Processing
  Letters}, vol. 23, no.2, February 2016, pp. 242--246.

\bibitem{pang17}
J.~Pang and G.~Cheung, ``Graph {Laplacian} regularization for image denoising:
  Analysis inn the continuous domain,'' in \emph{IEEE Transactions on Image
  Processing}, vol. 26, no.4, April 2017, pp. 1770--1785.

\bibitem{haynsworth68}
E.~V. Haynsworth and A.~M. Ostrowski, ``On the inertia of some classes of
  partitioned matrices,'' in \emph{Linear Algebra and its Applications}, vol.
  1, no.2, 1968, pp. 299--316.

\bibitem{bredies15}
K.~Bredies and M.~Holler, ``A {TGV}-based framework for variational image
  decompression, zooming and reconstruction. {Part I}: Analytics,'' in
  \emph{SIAM Jour}, vol. 8, no.4, 2015, pp. 2814--2850.

\bibitem{robbins12}
A.~Robbins and W.~Miller, \emph{Circuit Analysis: Theory and Practice}.\hskip
  1em plus 0.5em minus 0.4em\relax Cengage Learning, 2012.

\bibitem{daubechies10}
I.~Daubechies, R.~Devore, M.~Fornasier, and S.~Gunturk, ``Iteratively
  re-weighted least squares minimization for sparse recovery,'' in
  \emph{Communications on Pure and Applied Mathematics}, vol. 63, no.1, January
  2010, pp. 1--38.

\bibitem{RobustBoost09}
Y.~Freund, ``A more robust boosting algorithm,'' May 2009,
  https://arxiv.org/abs/0905.2138.

\bibitem{speriosu11}
M.~Speriosu, N.~Sudan, S.~Upadhyay, and J.~Baldridge, ``Twitter polarity
  classification with label propagation over lexical links and the follower
  graph,'' in \emph{Conference on Empirical Methods in Natural Language
  Processing}, Edinburgh, Scotland, July 2011.

\bibitem{wang15}
Y.~Wang and A.~Pal, ``Detecting emotions in social media: A constrained
  optimization approach,'' in \emph{Twenty-Fourh International Joint Conference
  on Artificial Intelligence}, Buenos Aires, Argentina, July 2015.

\bibitem{hu15}
W.~Hu, G.~Cheung, A.~Ortega, and O.~Au, ``Multi-resolution graph {Fourier}
  transform for compression of piecewise smooth images,'' in \emph{IEEE
  Transactions on Image Processing}, vol. 24, no.1, January 2015, pp. 419--433.

\bibitem{zeng16}
J.~Zeng, G.~Cheung, and A.~Ortega, ``Bipartite subgraph decomposition for
  critically sampled wavelet filterbanks on arbitrary graphs,'' in \emph{IEEE
  International Conference on Acoustics, Speech and Signal Processing},
  Shanghai, China, March 2016.

\bibitem{shomorony14}
H.~Shomorony and A.~S. Avestimehr, ``Sampling large data on graphs,'' in
  \emph{Symposium on Graph Signal Processing in IEEE Global Conference on
  Signal and Information Processing (GlobalSIP)}, Atlanta, GA, December 2014.

\bibitem{chen15sampling}
S.~Chen, R.~Varma, A.~Sandryhaila, and J.~Kovacevic, ``Discrete signal
  processing on graphs: Sampling theory,'' in \emph{IEEE Transactions on Signal
  Processing}, vol. 63, no.24, Cecember 2015, pp. 6510--6523.

\bibitem{knyazev15_mlsp}
A.~Knyazev and A.~Malyshev, ``Accelerated graph-based spectral polynomial
  filters,'' in \emph{IEEE International Workshop on Machine Learning for
  Signal Processing}, Boston, MA, September 2015.

\bibitem{knyazev15}
A.~Knyazev, ``Edge-enhancing filters with negative weights,'' in \emph{IEEE
  International Conference on Signal and Information Processing (GlobalSIP)},
  Orlando, Florida, December 2015.

\bibitem{gadde15_globalSIP}
A.~Gadde, A.~Knyazev, D.~Tian, and H.~Mansour, ``Guided signal reconstruction
  with application to image magnification,'' in \emph{IEEE International
  Conference on Signal and Information Processing (GlobalSIP)}, Orlando,
  Florida, December 2015.

\bibitem{zelazo14}
D.~Zelazo and M.~Burger, ``On the definiteness of the weighted laplacian and
  its connection to effective resistance,'' in \emph{53rd IEEE Conference on
  Decision and Control}, Los Angeles, CA, December 2014.

\bibitem{chen16}
Y.~Cheng, S.~Z. Khong, and T.~T. Georgiou, ``On the definiteness of graph
  laplacians with negative weights: Geometrical and passivity-based
  approaches,'' in \emph{2016 American Control Conference}, Boston, MA, July
  2016.

\bibitem{chu16}
L.~Chu \emph{et~al.}, ``Finding gangs in war from signed networks,'' in
  \emph{22nd ACM SIGKDD Conference on Knowledge Discovery and Data Mining}, San
  Francisco, CA, August 2016.

\bibitem{knyazev17arxiv}
A.~Knyazev, ``Signed {Laplacian} for spectral clustering revisited,'' January
  2017, https://arxiv.org/abs/1701.01394.

\bibitem{kunegis10}
J.~Kunegis, S.~Schmidt, A.~Lommatzsch, J.~Lerner, E.~D. Luca, and S.~Albayrak,
  ``Spectral analysis of signed graphs for clustering, prediction and
  visualization,'' in \emph{SIAM International Conference on Data Mining},
  Columbus, Ohio, May 2010.

\bibitem{biyikoglu05}
T.~Biyikoglu, J.~Leydold, and P.~F. Stadler, ``Nodal domain theorems and
  bipartite subgraphs,'' in \emph{Electronic Journal of Linear Algebra},
  vol.~13, November 2005, pp. 344--351.

\bibitem{sandryhaila13tsp}
A.~Sandryhaila and J.~Moura, ``Discrete signal processing on graphs,'' in
  \emph{IEEE Transactions on Signal Processing}, vol. 61, no.7, August 2013,
  pp. 1644--1656.

\bibitem{sandryhaila14}
------, ``Big data analysis with signal processing on graphs: Representation
  and processing of massive data sets with irregular structure,'' in \emph{IEEE
  Signal Processing Magazine}, vol. 31, no.5, August 2014, pp. 80--90.

\bibitem{chen14}
S.~Chen, A.~Sandryhaila, J.~M.~F. Moura, and J.~Kovacevic, ``Signal denoising
  on graphs via graph filtering,'' in \emph{IEEE Global Conference on Signal
  and Information Processing}, Austin, TX, December 2014.

\bibitem{rudin92}
L.~Rudin, S.~Osher, and E.~Fatemi, ``Nonlinear total variation based noise
  removal algorithms,'' in \emph{Physica D}, vol. 60, no.1-4, November 1992,
  pp. 259--268.

\bibitem{brew10}
A.~Brew, D.~Greene, and P.~Cunningham, ``The interaction between supervised
  learning and crowdsourcing,'' in \emph{Computational Social Science and the
  Wisdom of Crowds Workshop at NIPS}, Whistler, Canada, December 2010.

\bibitem{shi00}
J.~Shi and J.~Malik, ``Normalized cuts and image segmentation,'' in \emph{IEEE
  Transactions on Pattern Analysis and Machine Intelligence}, vol. 22, no.8,
  August 2000, pp. 888--905.

\bibitem{huang05}
J.~Z. Huang, M.~K. Ng, H.~Rong, and Z.~Li, ``Automated variable weighting in
  $k$-means type clustering,'' in \emph{IEEE Transactions on Pattern Analysis
  and Machine Intelligence}, vol. 27, no.5, May 2005, pp. 657--668.

\bibitem{weyl12}
H.~Weyl, ``Das asymptotische verteilungsgesetz der eigenwerte linearer
  partieller differentialgleichungen,'' in \emph{Math. Ann.}, vol.~71, 1912,
  pp. 441--479.

\bibitem{higham98}
N.~Higham and S.~H. Cheng, ``Modifying the inertia of matrices arising in
  optimization,'' in \emph{ELSEVIER Linear Algebra and its Applications}, vol.
  275-279, May 1998, pp. 261--279.

\bibitem{golub12}
G.~Golub and C.~F.~V. Loan, \emph{Matrix Computations (Johns Hopkins Studies in
  the Mathematical Sciences)}.\hskip 1em plus 0.5em minus 0.4em\relax Johns
  Hopkins University Press, 2012.

\bibitem{dorfler13}
F.~D{\"o}rfler and F.~Bullo, ``Kron reduction of graphs with applications to
  electrical networks,'' in \emph{IEEE Transactions on Circuits and Systems I:
  Regular Papers}, vol. 60, no.1, January 2013, pp. 150--163.

\bibitem{clr}
Cormen, Leiserson, and Rivest, \emph{Introduction to Algorithms}.\hskip 1em
  plus 0.5em minus 0.4em\relax McGraw Hill, 1986.

\bibitem{milanfar13}
P.~Milanfar, ``A tour of modern image filtering,'' in \emph{IEEE Signal
  Processing Magazine}, vol. 30, no.1, January 2013, pp. 106--128.

\bibitem{klein93}
D.~Klein and M.~Randic, ``Resistance distance,'' in \emph{Journal of
  Mathematical Chemistry}, vol. 12, no.1, December 1993, pp. 81--95.

\bibitem{keel09}
J.~A.-F. et~al., ``Keel: A software tool to assess evolutionary algorithms to
  data mining problems,'' in \emph{Soft Computing}, vol. 13, no.3, February
  2009, pp. 307--318.

\bibitem{uk09}
L.~Spacek, ``Face recognition data, university of essex, uk,''
  http://cswww.essex.ac.uk/mv/allfaces/faces94.html, Feb. 2007.

\bibitem{sun13}
W.~Sun, G.~Cheung, P.~Chou, D.~Florencio, C.~Zhang, and O.~Au,
  ``Rate-distortion optimized 3d reconstruction from noise-corrupted multiview
  depth videos,'' in \emph{IEEE International Conference on Multimedia and
  Expo}, San Jose, CA, July 2013.

\bibitem{chow70}
C.~Chow, ``On optimum recognition error and reject tradeoff,'' in \emph{IEEE
  Transactions on Information Theory}, vol. 16, no.1, September 1970, pp.
  41--46.

\bibitem{gewiki10}
\BIBentryALTinterwordspacing
Wikipedia, ``Gershgorin circle theorem,'' April 2016. [Online]. Available:
  \url{https://en.wikipedia.org/wiki/Gershgorin_circle_theorem}
\BIBentrySTDinterwordspacing

\end{thebibliography}

\end{document}